**Unravelling Responsibility for AI**

Zoe Porter*[a], Philippa Ryan[a], Phillip Morgan[b], Joanna Al-Qaddoumi[b], Bernard Twomey[a], Paul Noordhof[c], John McDermid[a] and Ibrahim Habli[a]

*corresponding author[1]

a Department of Computer Science, University of York, York, England, UK

b York Law School, University of York, York, England, UK

c Department of Philosophy, University of York, York, England, UK

## Abstract

It is widely acknowledged that we need to establish where responsibility lies for the outputs and impacts of AI-enabled systems. This is important to achieve justice and compensation for victims of AI harms, and to inform policy and engineering practice. But without a clear, thorough understanding of what 'responsibility' means, deliberations about where responsibility lies will be, at best, unfocused and incomplete and, at worst, misguided. Furthermore, AI-enabled systems exist within a wider ecosystem of actors, decisions, and governance structures, giving rise to complex networks of responsibility relations. To address these issues, this paper presents a conceptual framework of responsibility, accompanied with a graphical notation and general methodology for visualising these responsibility networks and for tracing different responsibility attributions for AI. Taking the three-part formulation 'Actor A is responsible for Occurrence O,' the framework unravels the concept of responsibility to clarify that there are different possibilities of *who* is responsible for AI, *senses* in which they are responsible, and *aspects of events* they are responsible for. The notation allows these permutations to be represented graphically. The methodology enables users to apply the framework to specific scenarios. The aim is to offer a foundation to support stakeholders from diverse disciplinary backgrounds to discuss and address complex responsibility questions in hypothesised and real-world cases involving AI. The work is illustrated by application to a fictitious scenario of a fatal collision between a crewless, AI-enabled maritime vessel in autonomous mode and a traditional, crewed vessel at sea.

## 1. Introduction

The importance of establishing where responsibility lies for AI-based systems has been emphasised in most sets of ethical principles proposed for AI (Jobin, Ienca & Vayena, 2019; Fjeld et al., 2020), by the OECD (OECD, 2019), and in UNESCO's global standard on the ethics of AI (UNESCO, 2021). It is also starting to be recognised in emerging AI legislation worldwide (European Commission, 2021; European Commission, 2022; Stanford University, 2023). The ability to attribute responsibility for the outputs and impacts of AI technologies, which are both influenced by and impact the wider ecosystem in which they are developed and deployed (Stahl, 2023), is important for several reasons. Thinking through consequences of decision-making, and learning from incidents and accidents, can inform AI policy and engineering practice to improve future outcomes. Determining the location of responsibility will also help to secure justice and compensation for victims of AI harms (Cane, 2002; Duff, 2007; Van de Poel, 2011; Hansson, 2022b).

Pinpointing where responsibility for AI lies is a journey into some unexplored terrains. AI-enabled systems have two features which makes them unlike typical tools and devices. They are often under-specified; that is, increasingly complex machine learning (ML) techniques, which have driven the performance of most of today's AI systems -- for example in image recognition, path planning, and navigation -- enable them to achieve goals without an explicit set of instructions for doing so (McDermid et al., 2024). This means there is greater uncertainty about their outputs and direct consequences in the real world than traditional software-enabled devices. In the paper, when we refer to 'AI-enabled,' we mean a subset of this, namely 'ML-enabled.' AI is also increasingly deployed in autonomous applications, where decision-making functions are delegated to the system (Burton et al., 2020) With the development of large language models (LLMs) based on transformer architectures, new horizons for autonomous systems are emerging. When given tool or service access, these models can enable `agentic AI systems', which can accomplish complex goals independently over long time horizons (Chan et al., 2024; Acharya, Kuppan & Divya, 2025).

The increasing sophistication of AI has prompted debates about attributing responsibility to AI-enabled systems themselves (Sebastián, 2021). The consensus is that these systems still lack the kind of agency necessary for moral or legal responsibility. But that does not simplify attributing responsibility to their human designers, makers and users. Responsibility frameworks are undergoing a process of adjustment, with solutions canvassed in the literature on how to attribute responsibility for the outputs and impacts of AI technologies when the normal necessary conditions for doing so, such as control and knowledge, are less obviously met by human actors across the AI lifecycle (Matthias, 2004; Porter et al., 2018; Hakli & Mäkelä, 2019; Himmelreich, 2019; Tigard, 2021; Goetze, 2022; Prifti, Stamhuis & Heine, 2022; Kiener, 2022; Morgan, 2023). Moreover, like other complex systems, the use of AI represents an exacerbated case of the 'problem of many hands' (Thompson, 1980; Nissenbaum, 1996; Braham & Van Hees, 2012; Van de Poel, Royakkers & Zwart, 2015; Cooper et al., 2022; Cobbe, Veale & Singh, 2023), whereby the sheer number of actors influencing a system's behaviour and its consequences makes tracing responsibility accurately or fairly to *individual* actors extremely challenging. Establishing where responsibility lies for the outputs and impacts of AI is therefore a multifaceted challenge, involving multiple actors, each of whom may be involved with different aspects of an overall event involving AI in different ways.

We believe that conceptual clarity is a necessary prerequisite for addressing responsibility questions precisely and thoroughly. Understanding the different senses or types of responsibility, and how and when these apply, can help to support discussants from multiple disciplinary backgrounds, including technical and engineering backgrounds, to reason effectively about responsibility for incidents, decisions, actions and omissions involving AI. This, in turn, can lead to focused interventions and improvements in design and engineering practice. There is related work in this field, with taxonomies of different types of responsibility given in (Davis, 2012; Van de Poel, Royakkers & Zwart, 2015; Van de Poel & Sand, 2021; Zerilli, 2021; Hansson, 2022b). What this paper adds to that related work is a graphical notation for modelling scenarios so that the responsibility relations within a wider AI ecosystem can be visualised and reflected upon, and a general process for establishing when the different senses or types of responsibility can be appropriately attributed to different actors. This helps to move the conceptual clarifications out of the realm of the theoretical and enables the conceptual framework to be practically operational for stakeholders debating responsibility attributions for AI. Another piece of related work is from the critical technology theorist Stahl (Stahl, 2023), who emphasises the need to take an ecosystem view of responsibility for AI, which recognises *"the complexity of the network of existing responsibilities."* This paper can support efforts in that direction, and enable the networked nature of responsibility relations to be modelled and perceived diagrammatically. Several of the core components outlined in Stahl's view of responsibility are elucidated in this paper's conceptual framework.

The paper is organised as follows. In Section 2 we present the conceptual framework, which is a systematic decomposition of the statement 'Actor A is responsible for Occurrence O'. At the heart of the conceptual framework is the 'unravelling' of responsibility into four senses; these are four different types of responsibility, or four different *ways of being responsible*: causal, role, liability, and moral responsibility. This four-fold distinction, like many responsibility taxonomies (Zerilli, 2021; Hansson, 2022b), derives from the legal philosopher H.L.A Hart (Hart, 2008). The informal graphical notation for different permutations of 'A is responsible for O' is presented, based on the work of Ryan et al. (Ryan et al., 2024), along with criteria and conditions for each of the four types of responsibility. In Section 3, we show how the criteria and conditions can be used to highlight inappropriate or unjust responsibility attributions, whether the overloading of responsibility on some actors, or the evasion of responsibility by others. In Section 4, we apply the framework by modelling the responsibility relations, using the graphical notation, in a hypothetical incident caused by an AI-enabled autonomous vessel in the maritime domain. As we work through the example, we propose a general method for applying the framework and for considering the various responsibility attributions systematically. In Section 5, general uses of the work are considered. There are several intended users: the safety engineering and safety science communities; responsible technology researchers; corporate officials; public officials and policy makers; and teachers and students.

## 2. Unravelling Responsibility: Actor A is responsible for Occurrence O

The concept of responsibility picks out a kind of *relation* between an actor and an occurrence (Shoemaker, 2015). Responsibility is ascribed *to* actors. Actors are responsible *for* occurrences. And there are different *ways* in which actors can be responsible for occurrences.

In this section, we present the conceptual framework, which is a systematic decomposition the three-part statement 'Actor A is responsible for Occurrence O'. In Section 2.1, three subcategories of actor are presented. These are the subjects of potential responsibility attributions. In Section 2.2, we describe four different senses of responsibility. These are the distinct (but often overlapping) ways in which actors may be responsible for occurrences. In Section 2.3, we present subcategories of occurrence. These are the objects of potential responsibility attributions or, in other ways, things that actors may be responsible for. There can be many different permutations of these three elements, and some permutations are conceptually invalid. For example, we take it that an AI-enabled system (subcategory of actor) cannot be responsible in the moral or legal sense for any occurrence. The conceptual framework is presented with an informal graphical notation, as well as a set of criteria or conditions for each type of responsibility in subsections 2.2.1 – 2.2.4. The idea is that the notation and conditions will enable connected responsibility relations for different aspects of complex events to be modelled, visualised, and reflected upon.

An additional relational element of responsibility, namely, the responsibility of actors *to* others (Coeckelbergh, 2020; Vallor & Ganesh, 2023) is not directly included in the framework - although it is implicit; for instance, a legal duty of care (a subset of role-responsibility) is owed to those who are affected by the duty holder's actions. Being responsible to others is also something felt keenly by those who embody the character traits, attitudes, and emotions of responsibility-as-a-virtue. This sense of responsibility, which is not included as one of the four senses in Section 2.2, is discussed further in connection with role-responsibility and moral responsibility.

### 2.1 Actor A

In the majority of cases and incidents involving AI, multiple actors will be involved. We delineate three kinds of Actor A: an AI-based system; an individual human; an institution. To give some examples, an *AI-based system* could be an AI-based radiology imaging system, an AI-enabled maritime vessel, or an autonomous vehicle. Relevant *individual humans* would include data scientists, designers, programmers, safety engineers, users, individual operators, public officials, and corporate managers. *Institutions* would include the software development companies, manufacturers, public service operators, and national or international regulators and certification bodies. The three subcategories of Actor A are shown in Figure 1.

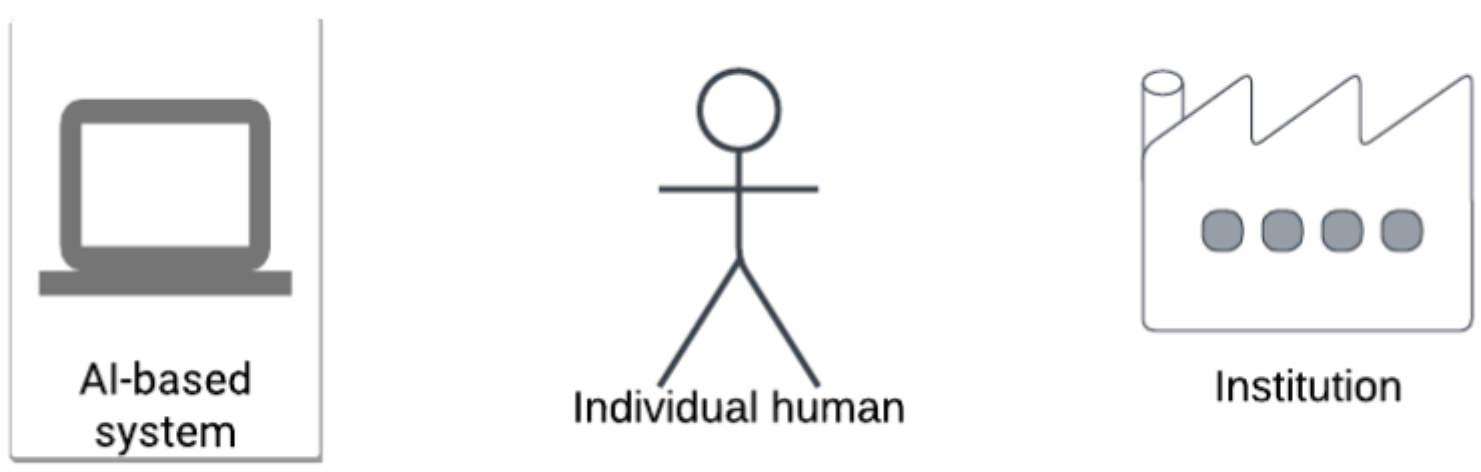

Fig 1: The three subcategories of Actor A

The three subcategories are not meant to be exhaustive. As our grasp of collective responsibility evolves, we might recognise non-institutional groups within the framework or view institutions as a specific type of collective entity with unique features. We might also consider new subcategories that, presently, would be created out of three core ones, for instance, human-AI teams (combinations of AI-enabled systems and individual humans), AI swarms (aggregates of AI-enabled systems or perhaps a single AI-enabled system taken as a whole), and sub-institutional groups (teams of individual humans). Similarly, *different kinds of relations between individuals* within and outside of institutions are not detailed here (for example, individuals may act together, or a few actors may act on behalf of the institution (Copp, 1979; Ludwig, 2017)). For usability, the framework is not presented at that level of granularity.

Not all kinds of actor can be responsible in the same way. More specifically, we assume that AI-based systems cannot be legally and morally responsible. AI systems are not legal persons in any legal system; thus, this subcategory of Actor A cannot have legal duties (see Figure 3) or be held liable (see Figure 4). In current legal scholarship, the standard position is to continue to advocate against the legal personhood of AI systems (Chesterman, 2020; Morgan, 2023). The dominant philosophical stance is that AI systems are not moral agents. As such, they also cannot have moral duties (see Figure 3) or be morally responsible (see Figure 5). While various reasons have been for this position, including their lack of sentience (Véliz, 2021; Verdicchio & Perrin, 2022), consciousness (Sebastián, 2021), understanding (Coeckelbergh, 2020), or the capacity to act from their own reasons (Johnson, 2006; Purves, Jenkins & Strawser, 2015; Verdicchio & Perrin, 2022), our justification is that AI systems are not moral agents because they are non-voluntary actors: they cannot choose not to act as they do.

By contrast, institutions are legally recognised as legal persons, thus this subcategory of Actor A can have legal duties (see Figure 3) and can be held liable (see Figure 4). Similarly, the prevailing philosophical view is that institutions are moral agents (French, 1984; Pettit, 2007; List & Pettit, 2011; Sepinwall, 2016; Björnsson & Hess, 2017), so we proceed on the basis that they can also have moral duties (see Figure 3) and be morally responsible (see Figure 5). We adopt a weak understanding of institutional moral agency as ultimately reducible to the moral agency of an institution's individual human members, who are voluntary actors, which we maintain AI systems are not. This simplification, while acknowledging the complexities of collective decision-making (Pettit, 2007; List & Pettit, 2011) suffices for the initial conceptual outline.

### 2.2 Is responsible for

Next comes the delineation of different ways of 'being responsible for'. Different actors may be responsible for different aspects of events involving AI in different ways. We refer here to different senses or types of responsibility. In common with related work in this space (Davis, 2012; Van de Poel, 2011; Van de Poel, Royakkers & Zwart, 2015; Zerilli, 2021; Hansson, 2022b), we draw upon H.L.A Hart's (Hart, 2008) taxonomy of the senses of responsibility. Getting clear on these different senses can help interlocutors from diverse disciplinary backgrounds avoid conceptual tangles. The four core senses are shown in Table 1.[2]

| Sense of responsibility | Description |
|---|---|
| Causal responsibility | A is a cause of O |
| Role-responsibility | A has tasks, moral duties or legal duties that attach to their role |
| Legal liability-responsibility | A is liable to legal sanction for O |
| Moral responsibility | A is an author of O (moral responsibility as attributability) |
| | A is liable to moral sanction for O (moral responsibility as accountability) |

Table 1: The main senses of 'is responsible for'

**2.2.1 Is causally responsible for.** The first and most general sense of responsibility is causal responsibility. We take 'causal responsibility' to be another way of referring to causality. We adopt a threshold notion of causality (Noordhof, 2020). That is, an actor needs only to have been *a* cause of O (not *the only* or *the most important cause* of O) to qualify as causally responsible. It is not only actors who can be causally responsible (Hart, 2008). Other occurrences O, facts, and events can also cause occurrences. The notation for an A's (and an O's) causal responsibility for an O is shown in Figure 2.

[2] Those familiar with Hart's taxonomy will note some differences between Table 1 and Hart's original presentation (Hart, 2008). Causal responsibility has been placed first rather than second in the list. This is to point to the fact that causal responsibility is the most general form of responsibility in the taxonomy. Another difference is that moral responsibility is expanded to include a distinction from Watson between moral responsibility as attributability and accountability (Watson, 1996). Finally, Hart's taxonomy also identifies 'capacity-responsibility' which we omit. 'Capacity-responsibility' is the possession of psychological capacities such as understanding, reasoning and self-control. It is a basic criterion for a *natural* person's being morally responsible and, in most cases, liable for what they do. In the framework, the capacity-responsibility of people (i.e., individual humans and the humans who comprise institutions) is assumed.

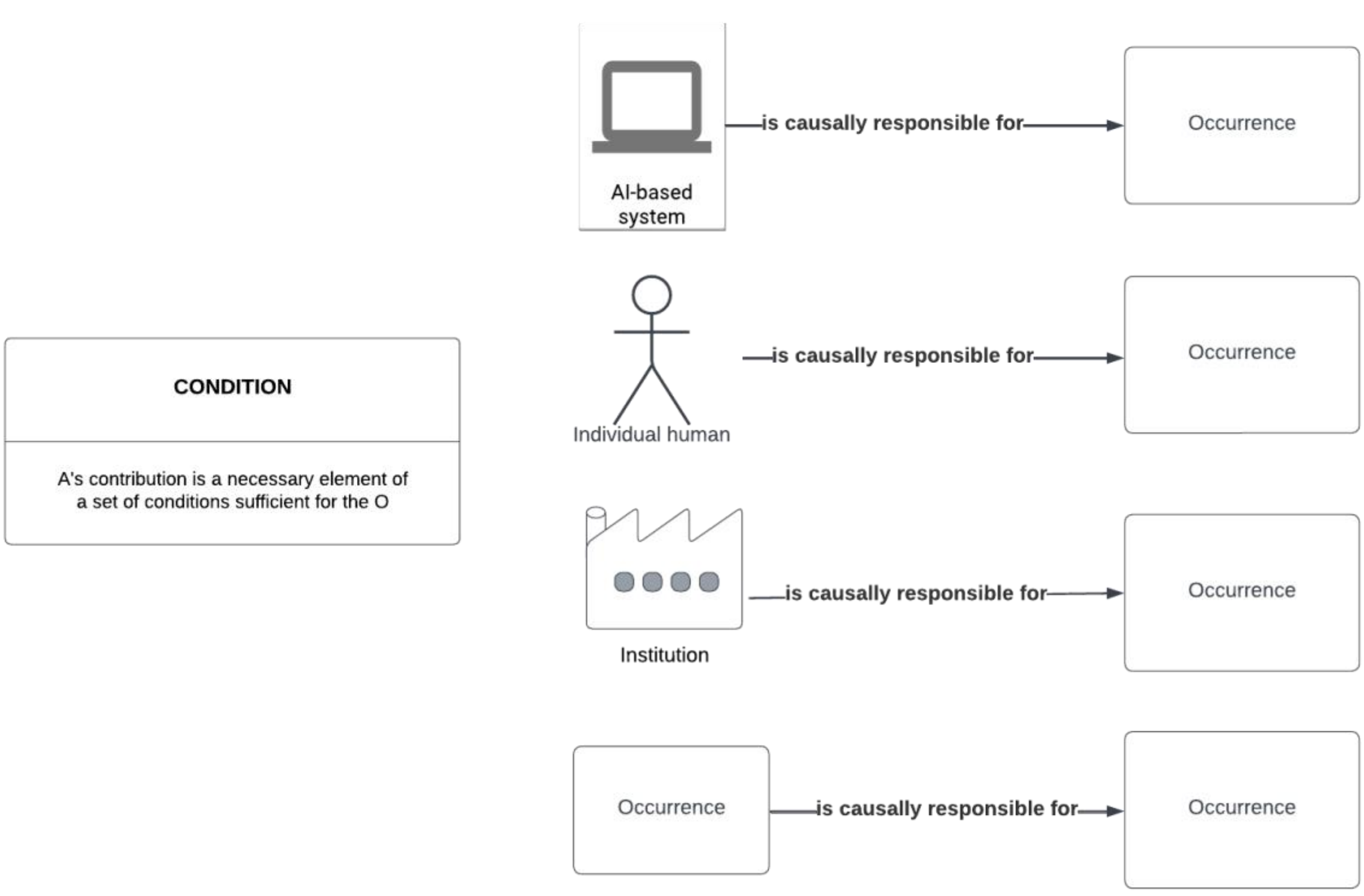

Fig 2: Actor A (or Occurrence O) is causally responsible for Occurrence O

By determining causal responsibility by reference to *a* cause rather than *the* cause of an occurrence, we can sidestep debates about what makes something *the most important* cause of an effect (Wittgenstein, 1976; Knobe, 2009; Danks, Rose & Machery, 2014; Hansson, 2022a; Tadros, 2018; Kaiserman, 2018). Many conditions for being a cause have been proposed in the literature (Beebee, Hitchcock & Menzies, 2009). To fix ideas, we suggest using the NESS condition, as shown in the box in Figure 2. The NESS condition is that a cause must be a Necessary Element of a Sufficient Set (Wright, 1985; Hart & Honoré, 1985). As Wright puts it, this is a *"test for causal contribution that is applicable to the entire spectrum of causation cases"* (Wright, 1985).[3]

Understanding and tracing causal responsibility is central to incident investigation and crucial when determining appropriate actions to mitigate future risk. Causation is therefore at the heart of safety engineering models, such as fault trees (U.S. Nuclear Regulatory Commission, 1981) and STAMP (System-Theoretic Accident Model and Processes) (Leveson, 2016). Understanding causal responsibility is part of the role-responsibility of safety engineers. Furthermore, causal responsibility is a necessary condition of moral responsibility and legal liability, and establishing causal responsibility relations can be an effective first step in attributing both. This general process is illustrated in Section 4, and the connections between different types of responsibility are detailed in Section 2.2.5.

[3] To deal with the case of effects of a common cause (where two or more things appear to be causally related but are in fact both caused by the same underlying factor), it is likely that appeal will need to be made to a more sophisticated framework, such as a counterfactual framework (which considers whether the effect would still have happened if the cause had not) or an interventionist framework (which considers whether changing the cause would change the effect) (Beebee, Hitchcock & Menzies, 2009; Pearl, 2009; Noordhof, 2020).

**2.2.2 Is role-responsible for.** The second sense of responsibility is role-responsibility. This is typically what is meant when people speak about having *responsibilities*. Role-responsibility refers to the tasks, duties, and functions attached to an actor's role in a group or society. We delineate three kinds of role-responsibility an actor might bear: *task responsibilities*; *moral duties*; and *legal duties*. To reiterate, as AI-based systems are neither legal persons nor, on our framework, moral agents, they cannot be bearers of moral or legal duties, they can only have tasks. The notation for an A's role-responsibility for an O is shown in Figure 3, along with a summary of the criteria for appropriate attributions of role-responsibility.

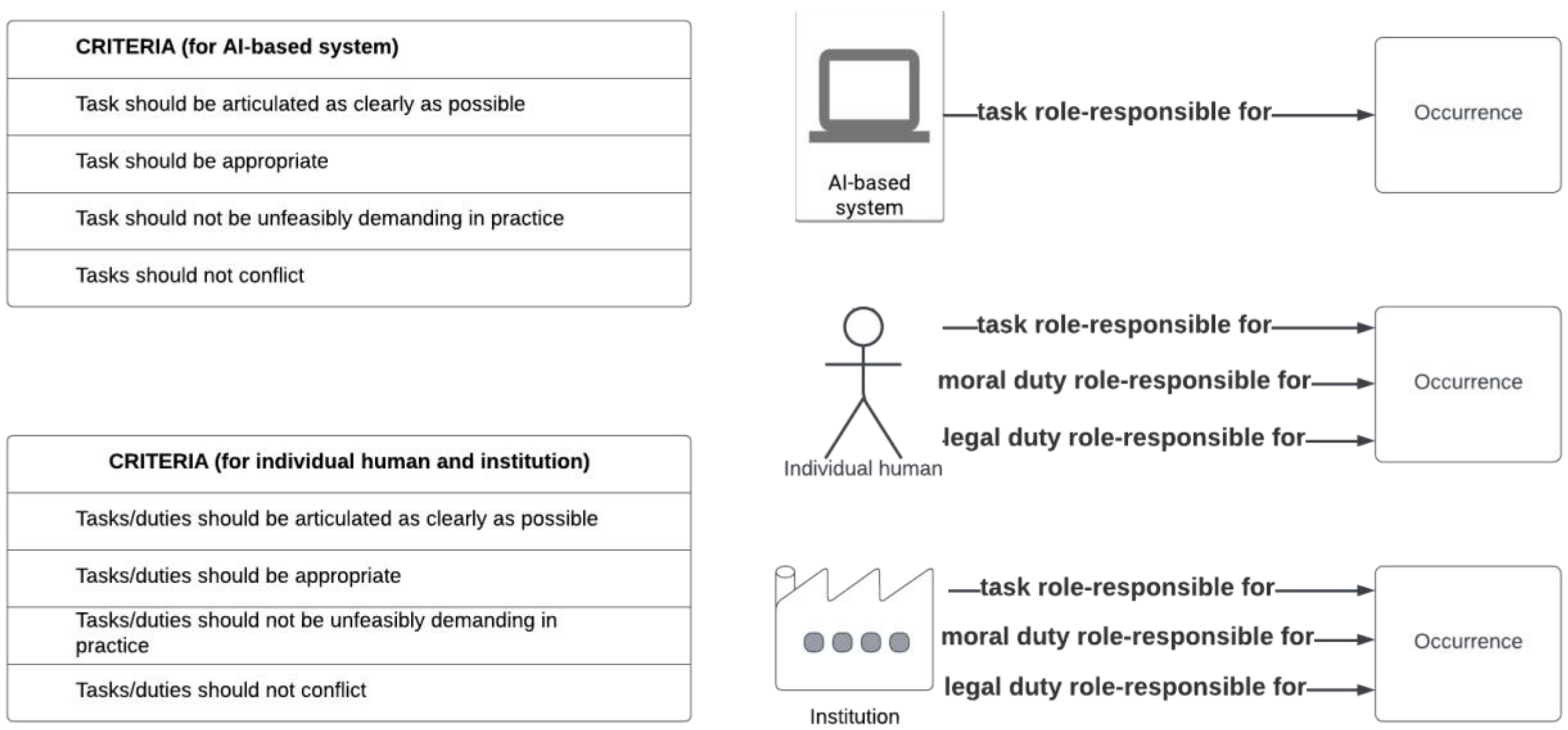

Fig 3: Actor A is role-responsible for Occurrence O

*Task responsibilities* are the jobs and functions assigned to an actor's role. Regarding AI, it is typically the engineer who defines an AI system's task responsibility, though responsible innovation encourages input from affected stakeholders during design (Ten Holter, 2022). *Moral duties* are moral requirements. Broadly speaking, we have a moral duty not to actively or unjustifiably harm others (because people have a right not to be harmed or wronged) and, in some circumstances, to come to the aid of others (Foot, 1967). The extent of actors' moral duties to protect users and those impacted by AI systems can be unclear. *Legal duties* are legally defined obligations, set out in law, such as the EU's AI Act (parts of which came into force in August 2024), with its duty on producers of high-risk AI system to carry out pre-deployment conformity assessments (European Parliament and Council, 2024). A legal duty may be a *duty of care* not to cause actionable harm through negligence (by failing to meet the standard of a reasonable person carrying out the same function) (Percy & Charlesworth, 2018) or an *absolute duty* that must be adhered to whatever the effort, time or cost, or another legal standard.

Legal duties are defined by the relevant laws. Beyond this, it is important to assess whether assigned role-responsibilities are appropriate. Unlike the other three senses of responsibility, for role-responsibility we have criteria (a benchmark against which the appropriateness of a role-responsibility attribution can be assessed) rather than conditions (requirements that must be met for the responsibility attribution to be true or just). We propose the criteria outlined in the boxes in Figure 3. First, the role-responsibility in question should be stated as clearly as possible. Ambiguously described role-responsibilities are difficult to perform. Second, the role-responsibility should be appropriate to the context. For instance, a system designer whose brief is to design a system which only needs to tolerate extreme cold should not be tasked to d effort expended on the task is proportionate to the need - an important for safety engineering, where effort should be focused on the most harmful risks. Third, the role-responsibility should be practically achievable given the actor's capabilities and available support. For example, expecting instantaneous human takeover of an autonomous vehicle is unrealistic (Law Commission and Scottish Law Commission, 2022). Fourth, the role-responsibility should not conflict with the actor's other tasks or duties. For example, an engineer should not be tasked to deliver the product on time if that creates conflict with their moral and legal duties to ensure that it is (acceptably) safe. This criterion addresses situations where an assigned task role-responsibility might require unethical or unlawful actions.

Often thought of as forwards-looking responsibility (Talbert, 2016), role-responsibility is connected to policies of Responsible Research and Innovation (Owen et al., 2013) and Responsible AI (Askell, Brundage & Hadfield, 2019; Dignum, 2019; Schiff et al., 2020; Okolo, 2023), which concern how actors should perform their roles to design, engineer, manufacture, use, and govern innovative technologies in ways that contribute to ethically and societally desirable ends and do not cause unjustified harm or compound existing inequalities (Von Schomberg, 2013; Okolo, 2023; Eke, Wakunuma & Akintoye, 2023). Understood as a virtue, responsibility refers to the moral character of the actor, including traits such as reliability, trustworthiness, care for others and future risk-bearers, integrity, moral imagination, and practical wisdom (Fahlquist, 2015; Van de Poel, Royakkers & Zwart, 2015; Van de Poel & Sand, 2021). Van de Poel & Sand defines the virtue of responsibility as a disposition to take responsibility Van de Poel & Sand, 2021); Williams defines it as the readiness to respond to a plurality of normative demands or, in other words, a readiness to balance our many obligations appropriately and ethically (Williams, 2008). Responsibility-as-virtue is crucial for active and responsive technology development (Van de Poel & Sand, 2021).

In many ways, responsibility-as-virtue (which is not included as one of the core senses in the framework) is connected to forwards-looking role-responsibility (Hart, 2008), and specifically to an actor's serious and diligent efforts to fulfil their moral duties or moral obligations. However, as Fahlquist argues, whereas role-responsibility is often more definable and assigned, responsibility-as-virtue is more open-ended and voluntarily assumed (Fahlquist, 2015) Actors who are responsible in this forwards-looking sense work hard to avoid risks to others and to reduce the harmful impact of their activities (Fahlquist, 2015). Vallor has proposed the cultivation of a kind of moral character that expresses 'technomoral' virtues (Vallor, 2016), which are mastered by promoting a relational understanding of moral obligations in technosocial contexts and paying habitual attention to the ethically salient features of situations within these contexts (Vallor, 2016; Farina et al., 2024). We envisage that this paper's framework, when used to model scenarios diagrammatically, could be a springboard for such exercises in moral imagination, and for thinking about the consequences of decisions and actions (Fahlquist, 2015).

**2.2.3 Is legal-liability responsible for.** The third sense of responsibility is legal liability. An actor who acts unlawfully is usually liable, according to other legal rules, to sanction, whether punishment or the payment of compensation (Hart, 2008). We divide legal liability-responsibility into criminal and civil liability only, as shown in Figure 4. For present purposes, public law, which regulates the behaviour of public bodies, is omitted from this paper's analysis. The possible permutations of 'A is liability-responsible for O' are restricted on two fronts. First, because AI cannot be liable for occurrences, A in these cases can only refer to individual humans and institutions. Second, since, to trigger liability, an (actionable) harm actually has to have occurred (liability is inherently a backwards-looking type of responsibility), the only subcategory of occurrence in Figure 4 is consequences (these are explained further in Section 2.3).

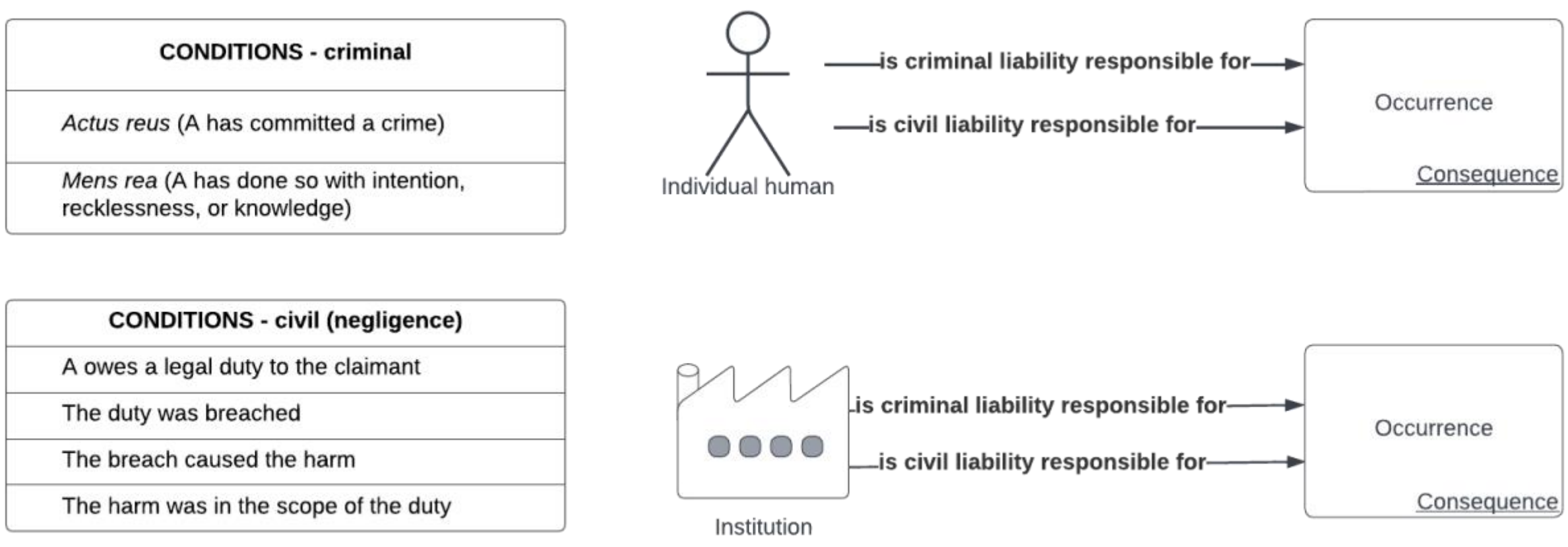

Fig 4: Actor A is legal liability-responsible for Occurrence O

The description of criminal and civil law below focuses on the Common law legal system. Common law legal systems include England and Wales, the United States, Australia, Ireland, New Zealand, Canada, India, Hong Kong, Singapore, Ghana, Uganda and Jamaica (amongst others). Many legal principles found within the common law are also found across developed legal systems, for instance in Civil law systems, many having a common taxonomic or historical root.

Criminal law concerns the prosecution of behaviour in the criminal courts. It primarily aims to safeguard the public and punish harmful acts, and is concerned with, amongst other things, the protection of bodily integrity, of property, and of the public welfare. In general, there are two conditions for criminal liability, which are summarised in the boxes in Figure 4. First, an action element (*actus reus*): the committing of the conduct necessary for a crime. Second, a mental element (*mens rea*): the doing so with the requisite intention, recklessness, or knowledge.

We have limited the decomposition of criminal liability to present a clear conceptual framework. Strict criminal liability and secondary liability could be added later as future developments, since these are particularly suited to considering criminal liability for AI. Strict criminal liability is often used in regulatory offences. It lacks the *mens rea* element. Secondary liability considers the aiding or encouragement of a principal offender in the completion of a criminal offence. The *mens rea* element is that they need to have had an intention to encourage or assist the offence.

Civil law regulates behaviour between parties. It seeks to determine the rights and duties of natural and legal persons, for example by establishing civil liability for a harm or wrong. The same conduct might concern more than one category of law (e.g., it might be both a criminal offence and a tort, a civil wrong). Conditions for civil liability also vary. In Figure 4 we pick out an area of civil liability that is particularly pertinent to cases involving AI. This is tort law (of which there are several subcategories) and, more specifically, negligence. Negligence is the workhorse tort in common law jurisdictions. Its elements, or conditions, are as follows: the actor owes a legal duty (of care) to the claimant; this duty was breached; the breach caused the claimant's harm; and the harm must be within the scope of the duty of care. These conditions are given in the civil liability box in Figure 4.

There are other, relevant types of tort liability which are not summarised in Figure 4. First, strict product liability (although defective products in some circumstances may also lead to liability in negligence). Second, vicarious liability, when an actor (typically an employer) is held civilly liable for the torts of another (typically their employee). This is not a separate tort but a way in which any of the torts can be attributed. In addition to tort law, contract law is a relevant branch of liability for AI-enabled systems, although contractual duties are only owed between contracting parties, and are not owed to third parties (except in very limited cases). Here, the condition is that an actor must have breached terms of their contract.

**2.2.4 Is morally responsible for.** The fourth sense of responsibility is moral responsibility. This is commonly thought of as a kind of backwards-looking responsibility (Van de Poel & Sand, 2021). Moral responsibility concerns the relation between an actor and an occurrence that calls for the actor to justify *why* they acted as they did. We draw a distinction between *being* morally responsible (moral attributability) and *being held* morally responsible (moral accountability) (Watson, 1996; Scanlon, 2000), as shown in Figure 5. There are more varieties of moral responsibility in the literature (Shoemaker, 2011), but we limit this framework to two subcategories for ease of use. Since AI-based systems cannot be morally responsible for an O, they are excluded from Figure 5.

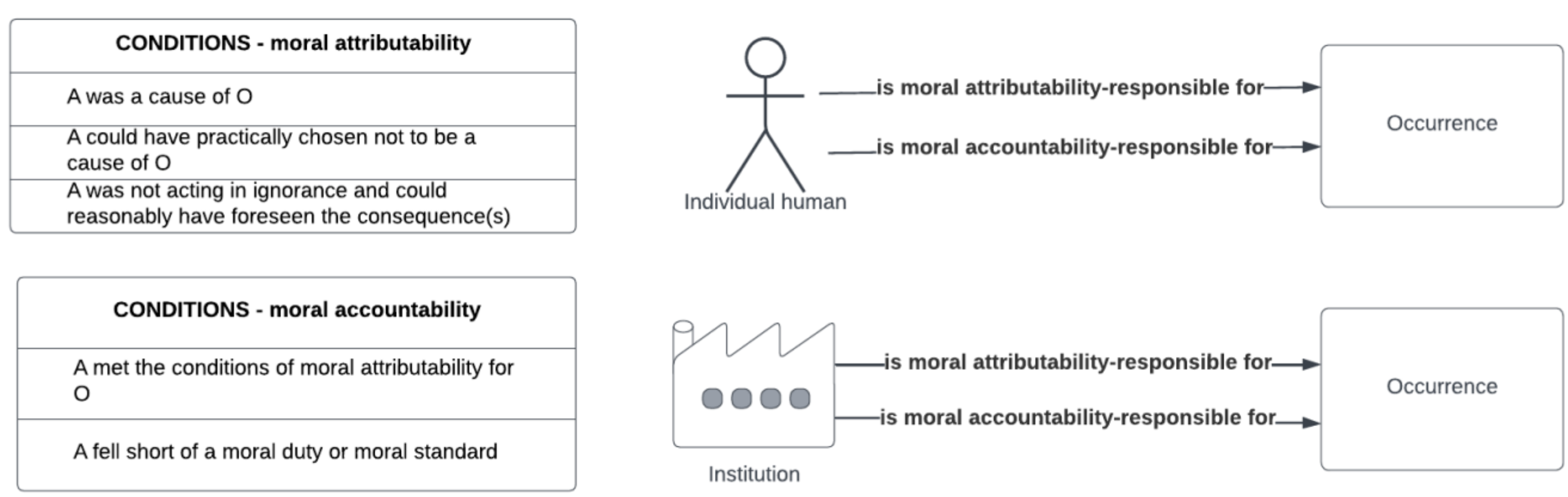

Fig 5: Actor A is morally responsible for O

Moral attributability picks out an actor's authorship of an occurrence. If O is morally attributable to Actor A, O was an exercise of A's voluntary agency. As such, it is reasonable to ask A for their reasons for performing or bringing about O (Porter et al., 2022). Doing so can prompt ethical reflection and contribute to better safety cultures (Dekker & Breakey, 2016). Drawing upon an Aristotelian tradition of moral responsibility, we propose the following conditions for moral responsibility as attributability: A was a cause of, or causally responsible for, O; A's causing O was an exercise of their voluntary agency, in that they could have practically chosen *not* to be a cause of O, for example because they were not under duress (control condition); A was not acting in ignorance and could have reasonably foreseen the consequences (knowledge condition) (Aristotle, 2011; Fischer & Ravizza, 1998; Robichaud & Wieland, 2017; Frankfurt, 2018). These conditions are summarised in the box in Figure 5.

Moral accountability concerns an actor's relations to other people (Watson, 1996). We take it that A is morally accountable when A deserves certain interpersonal responses such as resentment and indignation, leading to blaming practices like punishment and demands for apology (referred to in Table 1 as 'moral sanction') (Darwall, 2007; Strawson, 2008; McKenna, 2012). As a rule, A deserves these negative responses if A has failed to meet a standard of care and respect for others in the moral community (Strawson, 2008). We propose the following conditions for moral accountability, as shown in the box in Figure 5. First, A must meet the conditions of moral attributability for O, because there is an entailment relation between moral accountability and moral attributability (in other words, *being* morally responsible for O is a necessary condition for rightly *being held* morally responsible for O) (Watson, 1996; Shoemaker, 2013). Second, we propose that A must fall short of a moral duty or of a more general expected moral standard of care or respect for others (Strawson, 2008; Darwall, 2007). This reference to a standard of care represents an implicit connection to the notion of responsibility-as-virtue, discussed in Section 2.2.2. People who consistently meet or exceed standards of care and respect for others (Unruh, 2022; King, 2023) are responsible people in the virtue sense (Fahlquist, 2015).

**2.2.5 The relations between the four senses of responsibility.** The relations between the four senses of responsibility are summarised in Table 2.

| Sense of responsibility | Is necessary condition of |
|---|---|
| Causal | liability, attributability |
| Role-responsibility (legal duty) | liability |
| Role-responsibility (moral duty or moral standard) | moral accountability |

Table 2: The relations between the senses of responsibility

The fact that these are necessary not sufficient conditions is important to bear in mind. It helps to overcome some of the normative concerns that arise when people over-identify causation with liability and moral attributability. For example, in the world of safety science, `cause' is seen almost as a `trigger word' that should be avoided in an effective safety culture (Dekker & Breakey, 2016). In a recent NHS England document on patient safety incident investigations, for example, investigators are advised not to use the word `cause' at all, because it is *"strongly associated with blame and liability*" (NHS England, 2022). Remembering that causal responsibility is but a necessary condition of liability and moral responsibility helps to ensure that this worry is not overstated to the detriment of learning from the causal analysis of incident investigations. Causal responsibility is not a criterion of role-responsibility, but it is intimately connected to it. One way of thinking about this is to see role-responsibility as a kind of responsibility to be causally responsible for certain occurrences.

The recognition that role-responsibility is necessary but not sufficient for liability and moral accountability helps to constrain thinking that inquiries into liability and moral accountability are immediately answered once the relevant duty-holders have been identified. This is not the case: other conditions also need to be met. Furthermore, to be clear, with respect to moral accountability in Table 2, the claim is that in a system of appropriately assigned role-responsibilities, role-responsibility will be a necessary condition of moral accountability. There may have been a failure to assign the requisite moral duties to actors' roles, or the issue might be that the actor fell short of a generally expected moral standard of care and respect for others.

## 2.3 Occurrence O

In the project of locating responsibility for AI, it is important to be clear and specific about *what*, exactly, different actors are purported to be responsible *for*. To this end, we delineate seven different *types* of occurrence for which actors may be responsible, as presented in Figure 6.

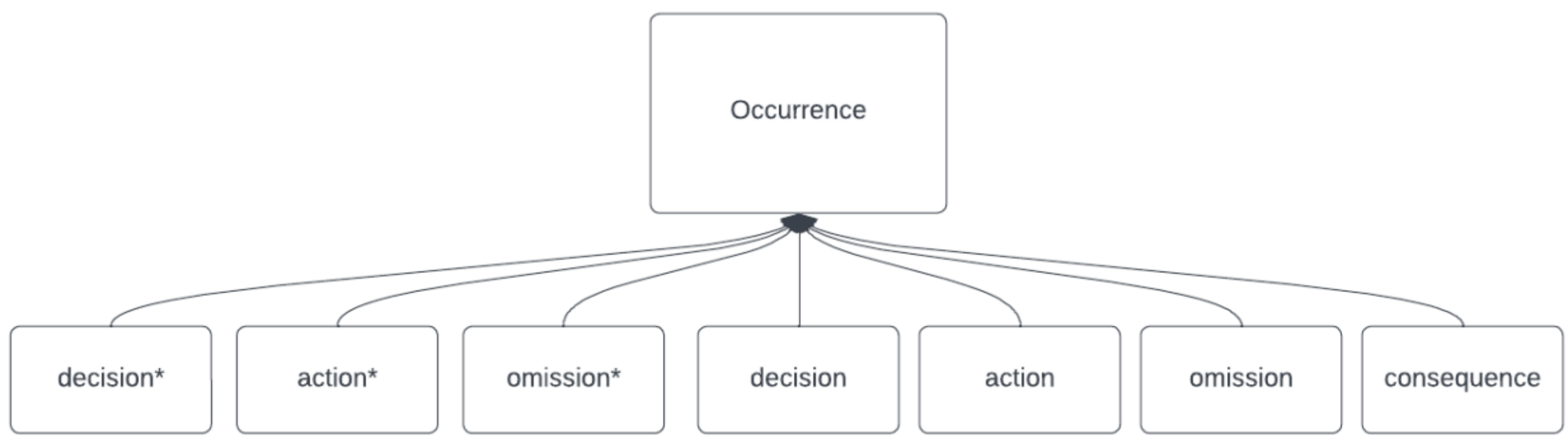

Fig 6: Subcategories of occurrence

Dividing 'O' in this way allows the framework to handle incidents involving multiple actors across different stages. This detailed breakdown may also aid safety analysis by scenarios to be modelled in which different occurrences cause system failure (Fenelon & McDermid, 1993; Carlson, 2012).

The first three subcategories in Figure 6 are specific to AI: *the outputs of AI-based systems*. Unlike traditional tools or devices, AI systems can reach outputs, including classifications, predictions, plans, and generated text or images, without being designed with a complete and explicit specification of how to do so (Burton et al., 2020; McDermid et al., 2024). We refer to these outputs as *decisions**. For cyber-physical AI systems, such as driverless cars and uncrewed vessels, their outputs also include sequences of physical manoeuvres; we refer to these outputs as *actions**. Increasingly autonomous systems can implement their *decisions** and *actions** without direct operational intervention. And we include *omissions** for cases where an AI system fails to generate an output. The star is included to distinguish the machine's outputs (or omissions) from a human's. This enables specificity in the modelling of complex cases. Since AI systems are not legal persons or moral agents, even where the *decision**, *action** or *omission** causes an incident, inquiries into liability and moral responsibility cannot stop at the AI system itself.

The second three subcategories in Figure 6 are the *decisions*, *actions* and *omissions* of individual humans and institutions. Individuals, albeit working in teams, make legion decisions and take legion actions related to the design, maintenance and use of AI-based technologies. They also make decisions about organisational cultures, system approval, and risk acceptance. Institutions make decisions and take actions in the form of board votes, corporate strategies and policies (Pettit, 2007; List & Pettit, 2011). Where it is unclear *which* individuals within an institution took decisions, actions or failed to act, it can be helpful in the first instance to attribute responsibility for these occurrences to 'institutions'.

Third, there are *consequences*. Consequences are the outcomes that are caused by any of the other subcategories of O as well as by physical events in the operating environment, such as storms and earthquakes. In inquiries of liability, it will be a consequence, such as a collision, injury, loss of life, damage to property or environmental damage for which liability will be sought, as illustrated in Figure 4.

## 3 Conceptual clarity to illuminate potential unjust responsibility attributions

We have set out the conceptual framework, providing the building blocks for articulating different possible permutations of A is responsible for O. We also presented the conditions for just attributions of responsibility. Clarity on these conditions, specifically the conditions of liability-responsibility and moral responsibility, can help users of the framework to illuminate and avoid unjust responsibility practices when applying it to specific scenarios.

### 3.1 Individual human actors as 'liability sinks' and 'moral crumple zones'

There is a documented concern in the automotive sector (Elish, 2019), in defence (Devitt, 2022), and in healthcare (Smith, 2021; Lawton et al., 2024) that, when AI-enabled systems cause or contribute to a harm or wrong, the nearest human operator will shoulder the burden of liability and moral accountability for bad outcomes. This is unfair. Just because they are the last in long causal chain of actors able to influence the impacts of AI-based systems, the nearest human operator -- the frontline clinician, the safety driver in an AV, the offshore remote operator -- is often not the most culpable for accidents and incidents.

In respect of liability, this has been called the 'liability sink' problem (Lawton et al., 2024) It arises because often the nearest human operator is the easiest actor to sue (Lawton et al., 2024). A real-world example is the sentence of endangerment of the safety driver in the Uber vehicle that killed Elaine Herzberg in Arizona in 2018. Though the safety driver had certainly been acting negligently, she alone carried criminal liability for the incident, despite well-documented engineering failures (National Transportation Safety Board, 2019) since because Uber ATG was not charged and she entered a guilty plea. A similar worry in the automotive domain arises in respect of drivers of automated vehicles who must accept, when activating `self-driving mode', that they will take over manual control immediately when required, even though in practice they often have insufficient time to do so effectively (National Highway Traffic Safety Administration, 2022).

In the medical sector, the liability sink is a particular concern in respect of AI-based decision-support tools. Here, the clinician is relied upon as a final 'safeguard' at the end of a long chain of design, development and deployment decision-making (Lawton et al., 2025). In such cases, if the clinician implements an incorrect AI recommendation on patient treatment or care, which may happen in complex cases (and where there is a wider cultural encouragement of deference to AI, which has been highlighted in this domain (Grote & Berens, 2020;  Tobia, Nielsen & Stremitzer, 2021)), the clinician risks absorbing all of the liability even though the system is at fault (Lawton et al., 2024; Lawton et al., 2025). For this reason, one study in the sector has advocated that healthcare systems and organisations should limit deployment to AI tools that simply present information and do not make direct recommendations until product liability for AI has been revised (Lawton et al., 2025).

In respect of moral responsibility, the concern is called the 'moral crumple zone' (Elish, 2019), because, like the crumple zone of a vehicle, the nearest human operator absorbs the shock of blame for wider system failure. This moral scapegoating may occur because their causal involvement, typically a failure to intervene in time or effectively, is the most obvious to observers. In addition, these individuals have a role-responsibility to be a final 'stop gap' before harm. It is then tempting and easy to say that, in cases where they fail to fulfil this role successfully, they (alone) should be blamed. But attending to the criteria of appropriate attributions of forwards-looking role-responsibility can help us to consider in specific whether too much is expected of the nearest human operator or they face a conflict in their roles (Crootof et al., 2023). Further, looking at the conditions of backwards-looking moral responsibility can help us to consider whether this individual could have practically chosen to act differently in the situation and whether they had adequate understanding of it. If not, then it would be unfair to attribute moral responsibility to them for the failure or harm.

Naturally, often the answers will be matters of degree. For example, it may be that the individual in question bears some of the fault but not all of it. The point is that the conceptual framework provides a language for identifying when individuals are being over-burdened and are at risk of bearing the brunt of legal and moral sanction downstream after an incident has occurred.

### 3.2 Powerful corporate and institutional actors evading or obscuring their responsibility

When frontline actors are singled out for sanction for adverse outcomes, it can obscure the truth that other individual and institutional actors upstream in the life cycle are equally and often more deserving of liability and moral accountability. The flipside of scapegoating is that other actors can evade legal or moral sanction.

By starting with actors bearing causal responsibility and role-responsibility for O and then considering whether they meet the conditions for liability and moral responsibility, we can illuminate where individual and institutional actors may also be at fault. For example, in the case of Uber Tempe, failing to build or deploy sufficient risk mitigation systems into the vehicle, which contributed to the incident, or failing to instil a good safety culture and employ engineers with appropriate safety knowledge (National Transportation Safety Board, 2019; Ryan et al., 2023).

Another way in which actors may seek to evade legal and moral sanction for harms caused or occasioned by AI is by hiding behind the complexity and independence of the AI-based system (Rubel, Castro & Pham, 2019; Cooper et al., 2022). The conceptual framework can also be used to illuminate this practice. For example, since AI-based systems are neither legal persons nor moral agents, any talk of *transferring* moral duties or moral responsibility to AI-based technologies could be identified as illegitimate. Indeed, such talk could be used by individual human and institutional actors to avoid their own moral duties or diminish their own moral responsibility for what occurs, shifting it onto AI-based systems (Stuart & Kneer, 2021). The framework can help interlocutors to illuminate such equivocation and obfuscation.

Unless we are to allow that responsibility can simply evaporate (Santoni de Sio & Van den Hoven, 2018), liability and moral responsibility for the outputs and impacts of AI systems must trace back to the individual human actors and institutional actors across the system's lifecycle and in the wider socio-technical system in which it is deployed - even though this undertaking is more complex in the case of AI systems than with traditional technologies.

# 4 Applying the conceptual analysis and graphical notation to model a responsibility scenario

We now demonstrate how the conceptual framework and notation can be made operational by applying it to a fictitious scenario involving an AI-enabled system in the maritime domain. As we work through the scenario, we describe a method for unravelling responsibility attributions within the network of responsibility relations. The aim is that this method helps to make the framework operational. It should be noted, however, that variations of this method could be supported by the framework. For instance, we start by establishing salient relations of causal responsibility, but another approach could be to start by establishing salient relations of role-responsibility.

This maritime example involves a cyber-physical AI-enabled system operating in an autonomous mode. It illustrates two aspects of AI that can complicate attributions of responsibility: underspecification in the ML pipeline (in this case the obstacle classification model); and the system's autonomous operation, which means that it can directly cause harm in the world without direct human involvement. The challenges these two aspects of AI can pose for attributing responsibility are discussed throughout the presentation of the example below.

Furthermore, this scenario illustrates the interconnected and networked nature of responsibility relations, and that several permutations of 'A is responsible for O' will be involved. Even when simplified to give an initial presentation of how the conceptual framework, notation, and method can be brought together to produce diagrams of responsibility attributions and prompt further reflection, the maritime example shows the multifaceted challenge of establishing where responsibility lies for AI.

For reasons of economy, we have not constructed illustrative examples involving other types of AI system, or other ethical complexities related to bias, explainability, and privacy. For an example of how the framework and notation presented in this paper has been applied in case involving the risk of bias from an ML model, please see Ryan et al. 2025 (Ryan et al., 2025).

### 4.1 Example case: fatal collision between autonomous ship and traditional crewed vessel

Imagine a crewless ship, which we will call a 'vessel'. For the most part, this vessel is controlled by a human operator from a shore-based remote operation centre (ROC). But it is capable of operating autonomously (i.e., without human intervention) and can do so where agreed with the regulatory authorities. In accordance with international best practice, the vessel and the whole maritime autonomous infrastructure are designed to be inherently safe and, where this is not possible, fault-tolerant.

This vessel is transporting cargo between Port A and Port B. It is being remotely controlled from the ROC by the remote human operator. The remote operator sees that, if the vessel's course is not altered, there will be a collision with another vessel. The second vessel is a traditional vessel crewed with STCW (Standards of Training, Certification and Watchkeeping for Seafarers) personnel. Before the remote operator can take action to adjust the remotely controlled crewless vessel's speed or course to avoid the collision, in accordance with the Convention on the International Regulations for Preventing Collisions at Sea (COLREGS) (International Maritime Organization, 2024), there is a loss of connectivity between the ROC and the crewless vessel. The remote operator therefore cannot send a control signal to the crewless vessel.

The vessel’s on-board systems detect the loss of connectivity and shift to autonomous operation. The vessel is fitted with an AI-based collision-avoidance system. The collision-avoidance system has several subsystems: it fuses data from multiple sensors on the vessel (cameras, lidar, radar and echo-sounders) to identify and classify obstacles in the area (both static, such as buoys, and dynamic, such as vessels); and it communicates with other on-board systems and functions to adjust course and/or speed to avoid any collision.

In our scenario, it is raining heavily and the camera lenses are covered in salt. This significantly degrades the quality of data from the cameras. As rain, and the subsequent salt build-up, is a foreseeable environmental condition, the cameras should be robust and well-maintained. In addition, the sensor-fusion subsystem, which uses a convolutional neural network (CNN), has been trained to classify obstacles, in various environmental conditions, when one of its input sources is compromised. Underspecification means that the model is likely to perform well on what it has been trained to classify (including within a certain range of image degradation) but it is unable to conclusively classify obstacles in edge cases (perhaps the extent of image degradation combined with the other vessel being positioned at an unusual angle in relation to the camera, or obscured by a wave). In the scenario, the extent of image degradation is particularly extreme because the sensors have not been sufficiently well-maintained. As such, on this occasion, the sensor-fusion subsystem fails conclusively to classify the obstacle in time which, in turn, means it fails to identify the course of action in time required to avoid a collision. Consequently, the ML-based collision-avoidance system does not communicate the correct inputs to the relevant other on-board systems to avoid the collision taking place.

The crew on board the other vessel expect the autonomous vessel to comply with COLREGS, which does not happen. Because it is now operating autonomously, is crewless, and has lost connectivity, the other vessel can only rely on a radio-based vessel-tracking system to get information about its position, course, and speed. But this tracking system does not convey information about the future intent of vessels. Once it is clear that the crewless vessel will not be altering course, the crew onboard the other vessel do not have enough time to take evasive action. The collision occurs, causing the loss of life of several members of the other vessel's crew. Significant environmental damage is also caused, due to a large spillage of fuel into the sea.

### 4.2 Modelling the scenario

Figure 7 presents the initial structure of the scenario using the graphical notation presented in Section 2. We work up from a basic model of the chain of events in the scenario in Figure 7 to a consideration of different types of responsibility attribution for different occurrences within this chain in Figures 8 – 11. As we do so, we suggest a general method for applying the framework.

The scenario as presented below concludes in a discussion of liability and moral responsibility. As such, it makes sense to start with one of the necessary conditions of these types of responsibility (causal or role) and build up from there. The method adopted here is to start with causal responsibility. The framework also supports looking at forwards-looking responsibility in greater depth, irrespective of later legal or moral sanction for any harms caused. Though we do not take such an approach in the example below, in such a case, it would make sense to start with role-responsibility, and to consider the extent to which roles for various occurrences have been assigned.

Before working through the different senses of responsibility and various permutations of 'A is responsible for O' in the maritime scenario, it is important to clarify that, although diverse responsibility attributions are being considered, we have kept the model relatively simple for an initial demonstration of how the framework can be applied. In reality, the distributed responsibility relations could become very complicated. Furthermore, responsibilities in real-world scenarios may be more ambiguous than they have been presented below. However, this does not undermine the value of this paper's framework. First, unless one is to write off the quest to attribute responsibility for AI as too complicated ever to pursue, one needs to start somewhere, and the framework and notation can provide a visual basis for doing so, as well as offer a structure for deeper deliberation. Second, the inclusion of criteria and conditions for each type or sense of responsibility can help users of the framework to articulate ambiguities precisely -- for example, by identifying where task-responsibilities have not been clearly enough articulated (role-responsibility), or by identifying where it remains unclear whether a harm was in the scope of a legal duty (liability). This clearer articulation can then inform and inspire interventions to prevent future ambiguities. Finally, the distillation of O into several subcategories can help to address some of the ambiguity because it can help to sharpen discussions on which bit of a wider picture an actor may be responsible for in any of the given senses.

### 4.3 Reasoning about causal, role, liability and moral responsibility for AI in the scenario

We now set out the method for unravelling responsibility by applying the framework and notation to the maritime example. First, we identify the main occurrence for which responsibility is sought. To manage complexity, we focus on the fatal collision (Consequence 2 in Figure 7) and not on the environmental damage. There are two pathways: the remote-controlled operation pathway and the autonomous operation pathway.

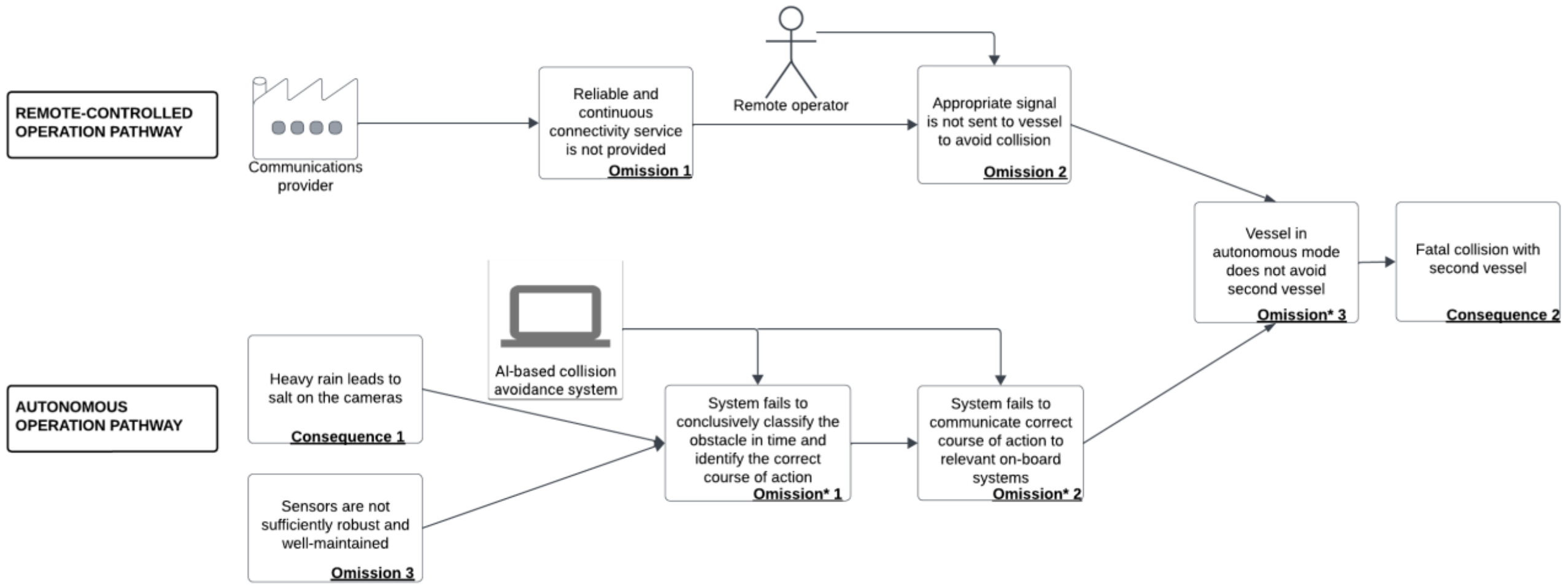

Fig 7: Initial model of the scenario using the notation

The approach then is to start with causal responsibility and then to layer up from there, adding role, liability and moral responsibility. As we do so, the initial model in Figure 7 is updated with additional actors and occurrences, and new responsibility relations between them.

**4.3.1 Causal responsibility.** Figure 8 expands on Figure 7 by picking out further relations of causal responsibility between actors (and occurrences) and other occurrences in the scenario. To note, we have omitted some variables from the causal model for the sake of clarity (Halpern & Hitchcock, 2020).

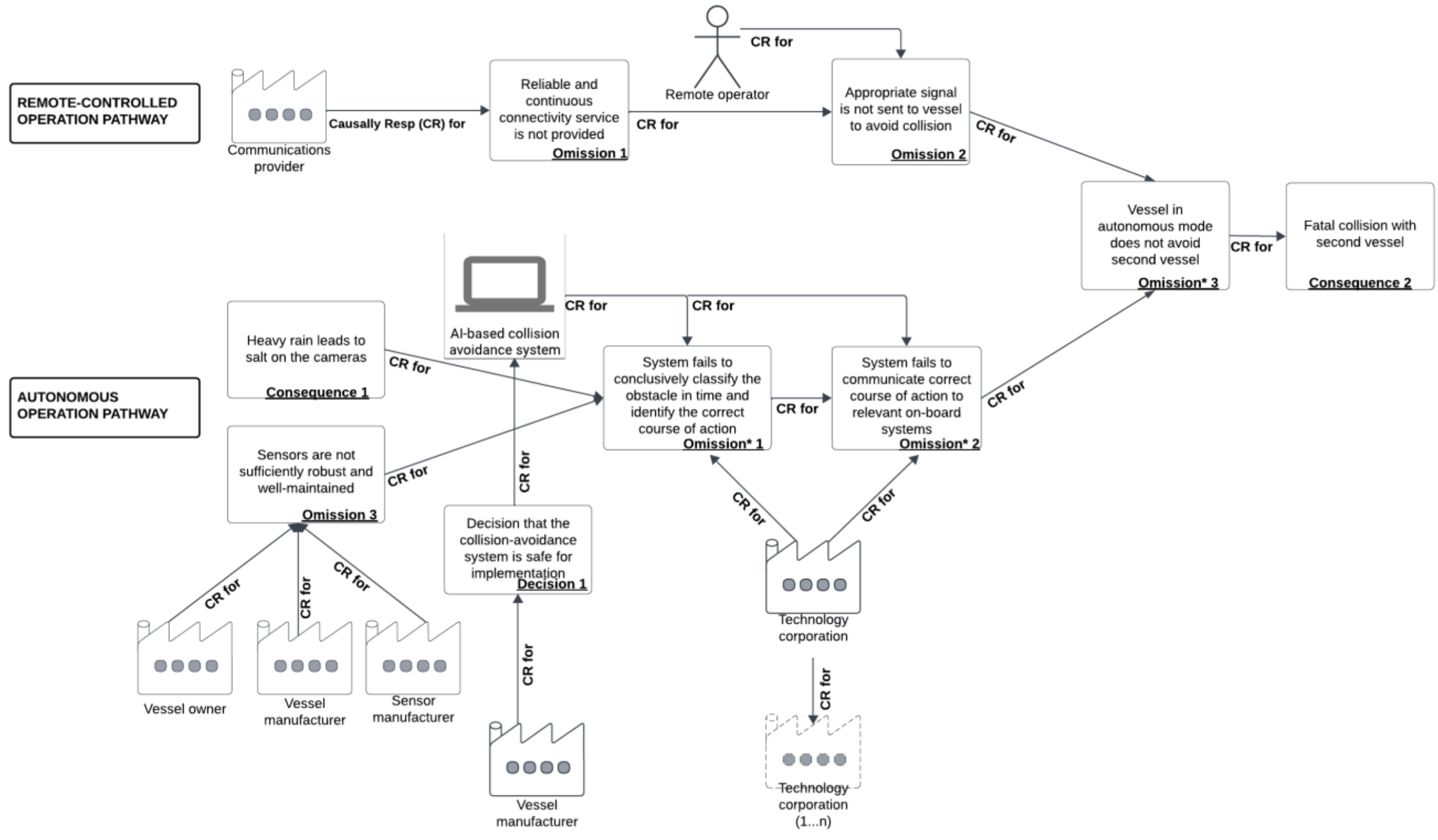

Fig 8: Modelling causal responsibility in the scenario using the notation

The figure does not identify any one cause as more salient to the collision (Consequence 2) than another. For example, the proximate cause of the collision (Omission* 3) is not identified as more significant than a distal cause (e.g. Decision 1). The framework could support more in-depth discussions about the relative weight of different actors' causal contributions. Several different measures for degrees of causal weight have been proposed in the philosophical literature (Tadros, 2018; Braham & Van Hees, 2009; Moore, 2009; Kaiserman, 2018; Demirtas, 2022), such as whether a cause constitutes a greater proportion of the outcome, or a greater proportion of the cause (Tadros, 2018). Questions about whether one cause, such as a 'root cause', is more significant to an outcome are also discussed in safety science (Del Frate et al., 2011). Representing scenarios diagrammatically as in Figure 8 can provide a useful focus for such discussions. And, because of this framework's inclusion of other types of responsibility, this approach can also help users to take that conversation further by considering whether different degrees of causal weight should influence degrees of liability or moral responsibility for the outcomes (Kaiserman, 2018; Tadros, 2018).

In our modelling of the maritime scenario, we use this backwards-looking model of causal responsibility as a starting point for considering where liability and moral responsibility may lie for the fatal collision. The notation could also be used for forwards-looking safety practices. The approach set out in Figure 8 bears some similarities with what is called a 'fault tree' in safety (U.S. Nuclear Regulatory Commission, 1981), which involves modelling undesirable events (such as omissions) to see how they could lead to a more consequential one. The framework can support also support forwards-looking effects analysis, to discover and better understand a failure mode and what its consequences might be, and to consider the appropriate corrective actions to manage future risk (Del Frate et al., 2011). This is particularly important in the case of AI, where failure modes are still poorly understood and evidence bases still lacking. The corrective actions deemed appropriate can then become explicitly articulated as part of actors' role-responsibilities in future cases.

**4.3.2 Role-responsibility.** Figure 9 expands on Figure 8 by picking out the role-responsibility relations between actors and occurrences in the scenario. It adds a new occurrence (the remote operator's role-responsibility for Action 1). It also adds one role-responsibility relation between actors (in the bottom right of Figure 9): the legal duty of technology corporations in the supply chain relied on by the main producer of the AI-based collision avoidance system. This legal duty would be a contractual obligation.

To reduce the number of actors in in Figure 9 for a simplified model, we assume that the vessel manufacturer has produced the entire ship, and that none of its sensors or shipboard functions, including the AI-based collision avoidance system, have been retrofitted. We also assume that the technology corporation that produces the AI-based collision avoidance system (which the vessel manufacturer configures, installs and validates in the vessels) has produced the system as a complete package intended for the maritime domain. The vessel manufacturer and the technology corporation would therefore both be deemed the 'producers' of the system.

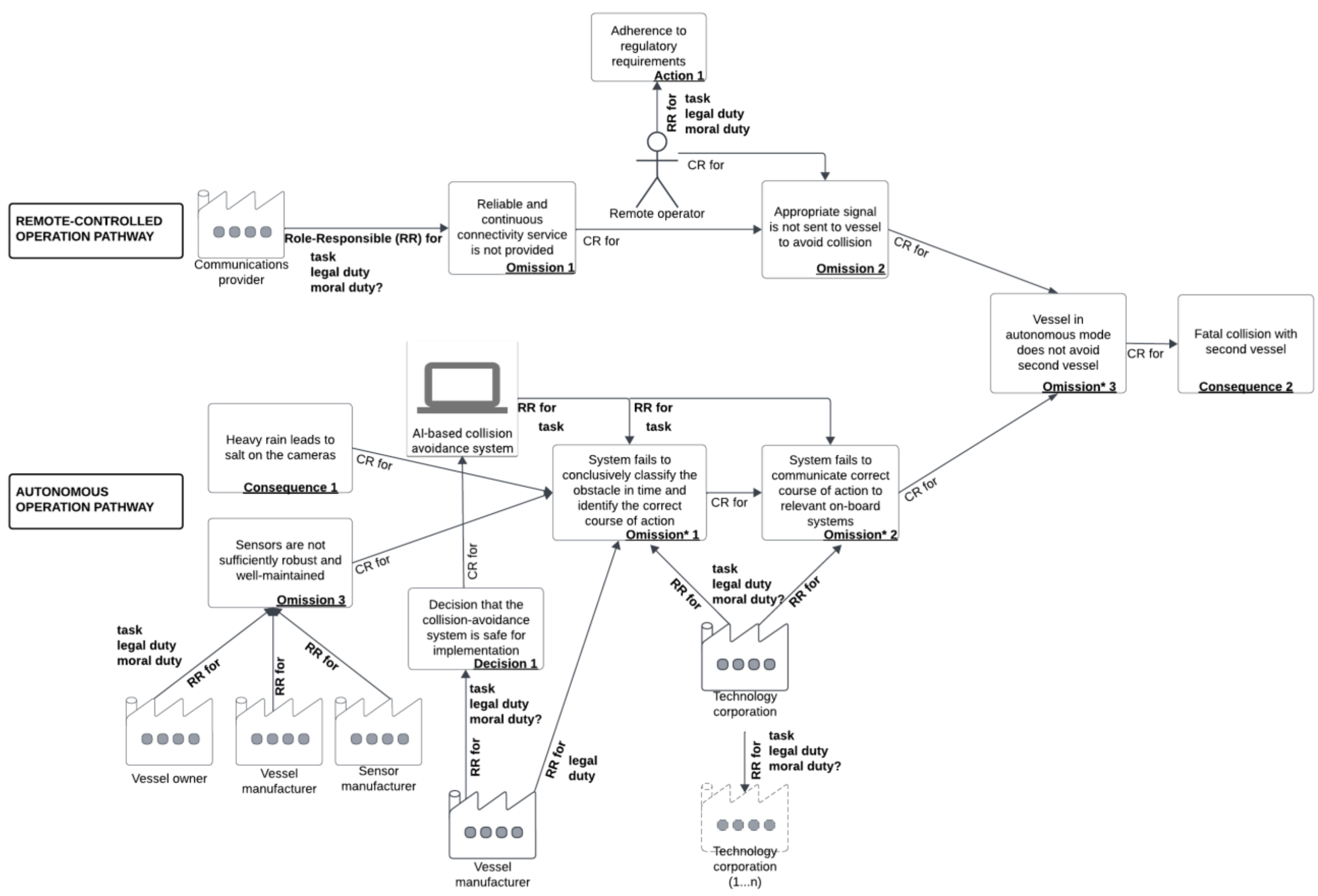

Fig 9: Modelling role-responsibility in the scenario using the notation

Because of space restrictions, where a role-responsibility exists in the same place as a causal responsibility in Figure 8, we have simply replaced the line with the label 'role-responsibility'. Role-responsibility implies one has a role to cause or prevent an occurrence (although one might not always fulfil one's role-responsibility). In a fuller and slightly larger depiction, however, different kinds of responsibility relation could be represented by separate, differently coloured lines on a single diagram. Where, in Figure 9, the line only states 'causally responsible for' (e.g., between Omission 1 and Omission 2), the notion is that this depicts causal but not role-responsibility.

Since the AI-based system (the collision avoidance system) is neither a legal person nor a moral agent, it only has task role-responsibility for Omission*1 and Omission*2 (or, more precisely, for performing those functions as intended). For the purposes of illustration, we have assumed that the communications provider, remote operator, vessel owner, vessel manufacturer, sensor manufacturer, and technology developing corporation all had legal duties to prevent the occurrences in the causal chain leading to the collision. The framework can support more fine-grained discussion about the content of legal duties are, and their regulatory or legislative source. The framework can also provide a structure for considering the nature and limits of these actors' moral duties not to harm people affected downstream by their decision-making and actions upstream. This can further be a springboard for considering responsibility-as-virtue and how the various actors involved would undertake their responsibilities if they were showing care and concern for future risk-bearers.

We have identified under-specification as a feature of AI which introduces novel questions for responsibility attribution for AI (Burton et al., 2020). In terms of role-responsibility, the questions here are as follows. When would we say that engineers at the vessel manufacturer (which, we have assumed, has produced the AI-based collision avoidance system) have fulfilled their role to make the system as safe as possible, given that underspecification makes such systems inherently uncertain and unable to classify all obstacles correctly? When would be able to say that relevant actors at the vessel manufacturer had sufficiently fulfilled their role-responsibilities when making Decision 1? What, precisely, would be the articulation of those role-responsibilities? Our purpose is not to answer these questions here, but to show how applying the framework diagrammatically can focus reflection on specific role-responsibility questions that arise due to the distinctive features of AI. We have also identified system autonomy as a feature of many AI systems that challenges attributions of responsibility. The maritime vessel is operating in autonomous mode. Amongst the questions here is the question for regulators (which are not depicted in Figure 9) about their role-responsibilities to clarify levels of uncertainty acceptance given the underspecification of AI-enabled systems, and their operational autonomy in real-world environments.

Once the role-responsibility relations of a scenario have been modelled in a similar way to Figure 9, users can cross-reference different actors' role-responsibilities with the criteria given in Section 2.2.2 to evaluate the clarity, appropriateness, demandingness and potential conflicts of tasks or duties. Contextually inappropriate, heavily demanding, or conflicting role-responsibilities could be made explicit (Bainbridge, 1983; Crootof et al., 2023; Endsley, 2023) For backwards-looking modelling, particularly of moral responsibility which will be discussed below, looking at back at these criteria later on can help to identify where some actors could not have practically chosen otherwise (e.g., because their task was too demanding) and therefore does not meet the control condition of moral attributability. For forwards-looking modelling, these criteria can highlight issues that need to be resolved for more responsible AI ecosystems.

**4.3.3 Liability.** Figure 10 expands on Figure 9 by picking out potential lines of liability. There are several interests at stake (property interests, insurer interests, cargo interests, and interests connected to the environmental damage) for which claims could be made, along with potential criminal prosecution for pollution offences. But for present purposes, we focus on the deaths of the seafarers. The initial modelling of liability for Consequence 2 is shown in Figure 10.

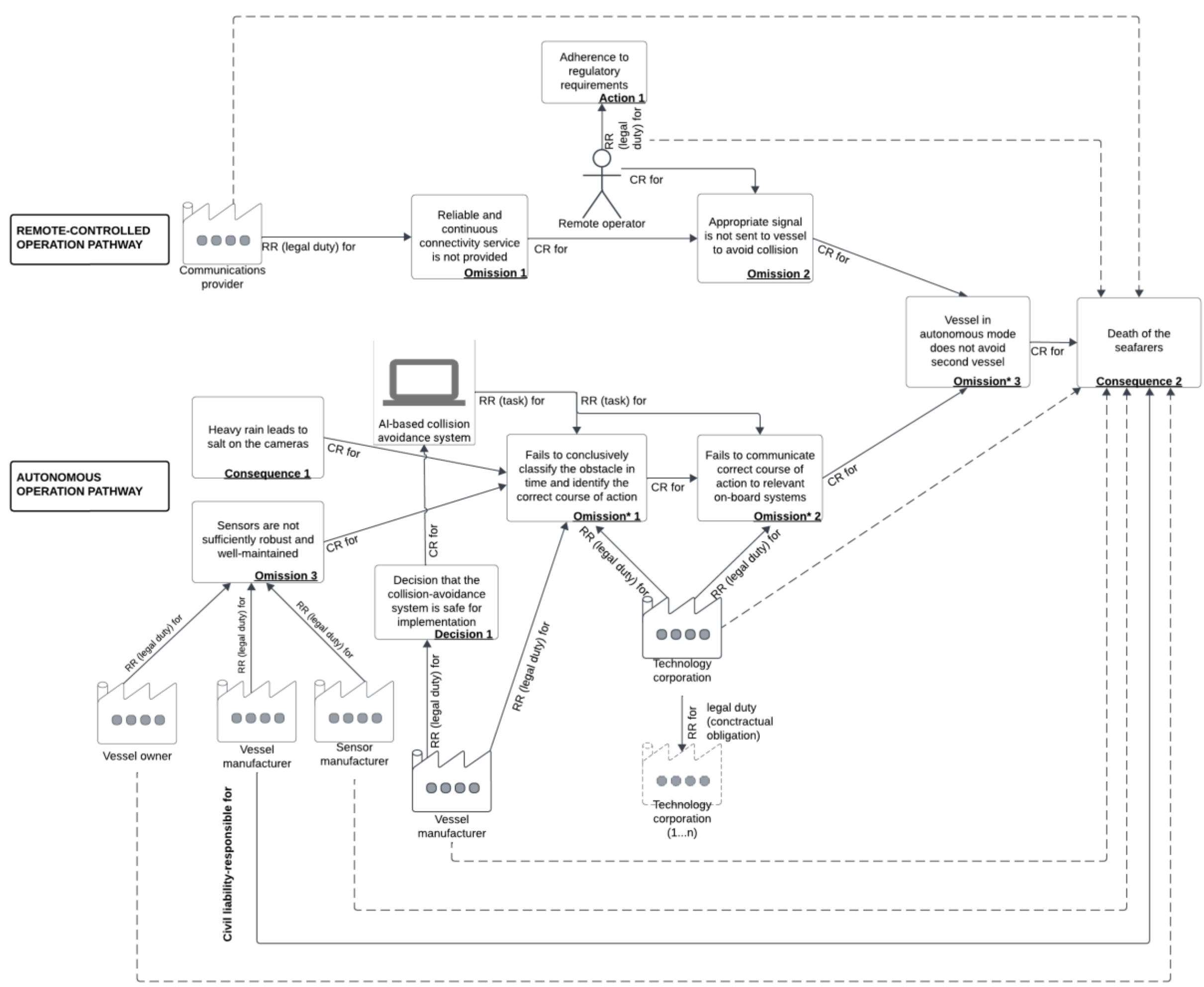

Fig 10: Modelling liability-responsibility in the scenario using the notation

The method for unravelling liability can proceed as follows. To unravel civil liability for the deaths, it needs first to be established where legal duties lie, and whether any of these were breached. For this, we can refer back to the role-responsibility in Figure 9 (for ease of reference, legal duties are also retained in Figure 10). We have hypothesised that the following actors have a legal duty: the communications provider; the remote operator; the vessel owner; the vessel manufacturer (which corporation we assume, to manage complexity in the example, configured, assembled and tested the vessel as a complete package, including with its sensors and AI-enabled functions, and is therefore a producer); the sensor manufacturer; and the technology corporation which provided the AI-based collision avoidance system (and is also a producer). There will be several technology corporations supplying elements of this system; these may have legal duties (in the form of contractual obligations). For present purposes, these will not be considered in the discussion.

Drawing on the conditions of civil liability, we might consider whether the actors each met their legal duties, or whether the omission in question reflected a breach of their duty, which then caused the harm, and the harm was in the scope of the duty. For example, were the vessel owners (of which there are, in practice, multiple layers of institutions) meeting their duty to ensure adequate inspection, maintenance and concept of operation of the cameras (Omission 3)? Did the vessel manufacturer who installed them do so correctly, or provide proper instruction as to their maintenance (Omission 3)? Did the sensor manufacturer consider the maritime environment, where they would be regularly exposed to salt water, and salt deposits (Omission 3)? Did the communications provider meet its duties of adequately maintaining the service, or supplying appropriate hardware or a service of satisfactory quality (Omission 1)? Regarding the AI-based collision avoidance system, did the technology corporation which engineered it, for installation by the vessel manufacturer, meet their duties to do so properly, and consider the foreseeable environmental conditions in which the system would need to work (Omission*1 and Omission*2)? Given the many inputs and parties involved, untangling and tracing the collision avoidance system's omissions* back to the actions of a legal or natural person may be extremely complicated (Morgan, 2024). In such cases, a claimant may therefore be tempted to pursue those involved in the design, manufacture, or maintenance of the cameras, since establishing this claim will be significantly easier. In Figure 10, we assume that the lawyer pursuing the civil claim has been successful in bringing a product liability claim against the vessel manufacturer either in negligence, or via a statutory product liability claim. This is denoted by the solid arrow in the figure from the vessel manufacturer to Consequence 2.

From questions of civil liability, we can also consider possible criminal liability. If the duties of care have been breached, one or more of these actors could face criminal prosecution, for instance via a charge of gross negligence manslaughter or corporate manslaughter (in the case of institutions), provided that the nature of the defendant's breach of its duty of care owed to the victim amounts to a gross breach of its duty. For these two offences the *men rea* element is typically considered to be a high degree of negligence. In addition, there may be offences committed under the Merchant Shipping Act 1995, for instance the potential for owner liability under Section 98 (dangerously unsafe ship); or the potential for owner, charterer, or manager liability under Section 100 (unsafe operation of ship). Further, the ship's failure to comply with COLREGS may result in the owner or any person 'responsible' for the conduct of the ship, being criminally liable under Regulation 6 of The Merchant Shipping (Distress Signals and Prevention of Collisions) Regulations 1996, unless they can show that they ‘took all reasonable precautions to avoid the commission of the offence.' The latter is a strict liability regulatory offence, with no *mens rea* requirement, but subject to a defence.

In Figure 10, potential lines of liability that could have been or can be traced back to other actors are identified with a dashed line. The model is partial because unravelling all the claims is beyond the scope of this paper. The aim is to show how the conceptual framework and graphical notation can be made operational to consider and communicate possible liability for AI-occasioned harms. As with the figures for the other senses of responsibility, this approach can also be a springboard for deeper discussions, such as the validity and feasibility of pursuing other civil claims or criminal prosecution.

**4.3.4 Moral responsibility.** Finally, Figure 11 adds moral responsibility attributions to actors for occurrences in the scenario. There is no specific need to model moral responsibility after, rather than before, liability -- although many actors have a greater interest in knowing what they would be liable for than what they would be morally responsible for, which might be one reason to consider liability first. To note, Figure 11 is, like Figure 10, a partial model. Here, too, we limit the number of potential attributions of moral responsibility that are discussed in detail. But the framework does provide a basis for such discussions.

We suggest the following method for unravelling moral responsibility. First, to unravel the moral attributability, users of the framework could, in the first instance, consider whether the conditions for moral attributability are met by the identified actors in the scenario. To recall, the conditions were that the actor needs to have been a cause of the occurrence, have practically chosen not to be so (control condition), and was not acting in ignorance and could reasonably have foreseen the consequences (knowledge condition). To be clear, to limit scope we limit our discussion to moral attributability for the fatal collision (Consequence 2) rather than any of the other occurrences in the scenario.

Let us imagine that the cause condition is met by these actors as depicted in the causal responsibility model in Figure 8. Users of the framework could then consider whether these actors met the control condition. For example, they could ask whether it was practically possible for the actor in question, say the vessel owner, to have chosen to ensure that the cameras were better maintained (Omission 3). To give another example, they could ask whether the vessel manufacturer (or individuals within that corporation) could have chosen not to decide that the collision-avoidance system was safe for implementation (Decision 1). In the latter case, we might ask whether any individuals there were acting under duress. It seems clear that the remote operator's failure to send a control signal to the autonomous vessel (Omission 2) was not something they could practically control, since there was a loss of connectivity with the vessel. As such, Omission 2 is not morally attributable to the remote operator. We might, however, question whether there was anything else the remote operator could have done, such as raise an alarm. This requires a new occurrence (i.e., Action 2 in Figure 11) in the model. Imagine the remote operator did not raise an alarm to the other, crewed vessel when they could have. Imagine that they breached a moral duty in failing to do so. This possibility is shown in the dashed line from the remote operator to the consequence in Figure 11.

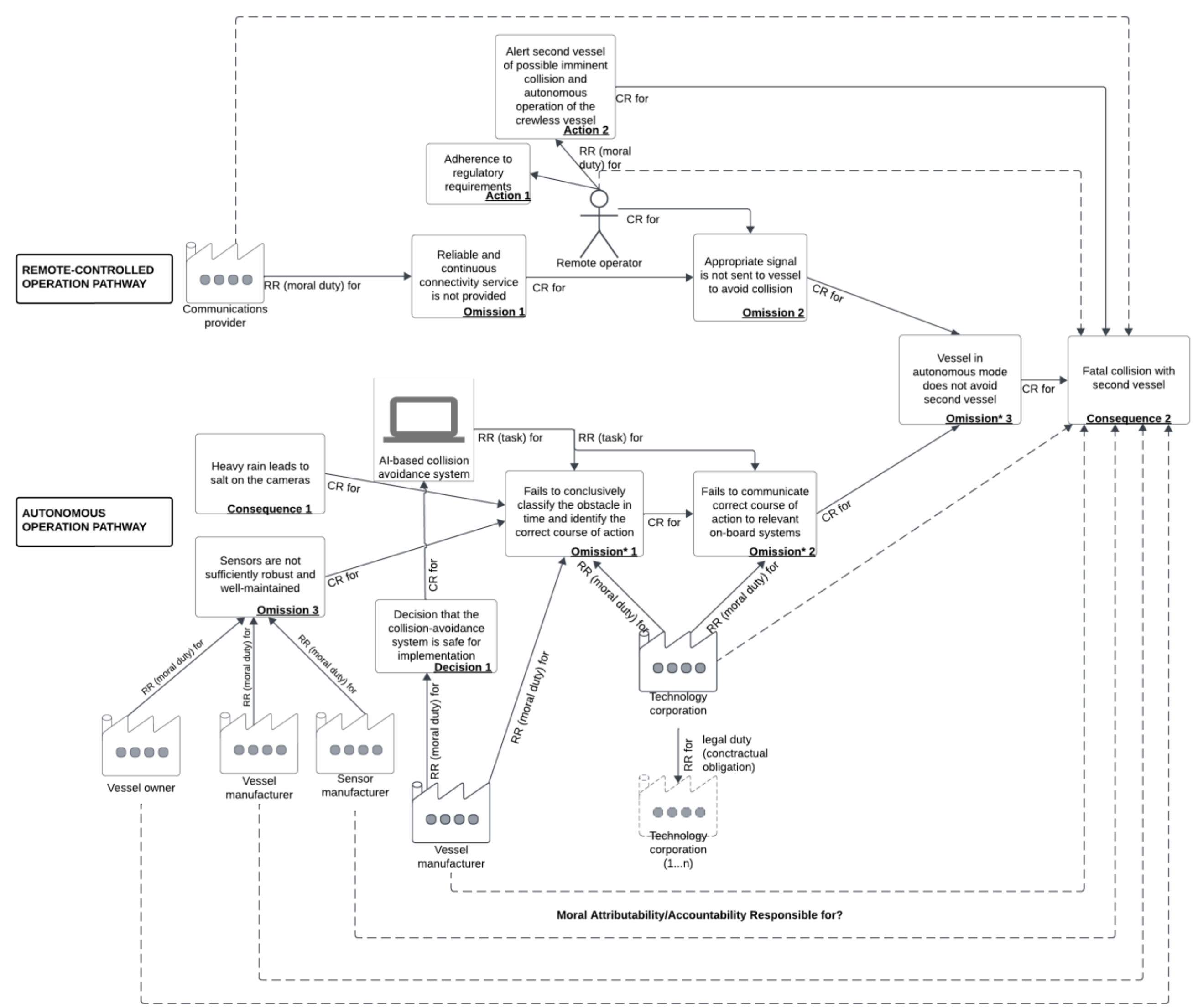

Fig 11: Modelling moral responsibility in the scenario using the notation

Next, users of the framework could consider how far these actors met the knowledge condition. Picking up on the same two actors focused on above, it seems likely that the vessel owner would have foreseen that inadequate maintenance of the cameras (Omission 3) raised the risk of a collision (Consequence 2) when the vessel was operating in autonomous mode. And with respect to Decision 1 and the causal chain to Omission*1, we might ask whether the vessel manufacturer could have foreseen that the underspecification of the collision-avoidance system and its inevitable inability to classify all obstacles correctly was a serious enough safety problem that could lead to a fatal collision (Consequence 2)?

There are further possible attributions of moral attributability that could be discussed. In the interests of economy, we do not undertake that discussion here, but possibilities are shown in the dashed lines between various institutional actors and the consequence in Figure 11. It is important to recognise that following this method will likely reveal that moral responsibility as attributability is distributed across many actors in the network of actors depicted in the scenario. Just picking out the two actors discussed above, we can see that they plausibly meet the conditions of moral attributability for the fatal collision itself. The value of the framework and notation is that it can enable a clear visual representation of this complex web of distributed moral responsibility, and thereby to provide a springboard for deeper discussions about whether moral attributability should be weighted with a smaller number of key actors, perhaps because they had more control than others in the responsibility ecosystem.

After considering moral attributability, users of the framework might then consider which of the actors would justifiably be morally accountable for the fatal collision. For moral accountability, the actor needs to have met the conditions of moral attributability and have fallen short of moral duty or standard. The 'standard' is included to refer to a more general expected behaviour in a moral community or society that is not specifically attached to a role as is a 'moral duty'. Thinking about whether decision-makers upstream fell short of a moral duty or an expected standard of care and regard for the risk-exposed can illuminate where there are potential unjust asymmetries in the weight of moral accountability borne by individuals and institutional and corporate actors. Different candidates who may be assessed for moral accountability for the deaths of the seafarers are identified by the arrows from these actors to the consequence in Figure 11. Again, the distributed nature of moral accountability for the harmful consequences of AI will be revealed by applying the framework and notation.

## 5 Discussion

Motivated by the belief that conceptual clarity is necessary to reason effectively about the multi-faceted challenge of attributing responsibility for the outputs and impacts of AI, this paper has presented an unravelling of the concept of responsibility based on the central observation that there are several possible permutations of 'Actor A is responsible for Occurrence O'. To ensure that this framework is practically operational for stakeholders deliberating about responsibility, including practitioners, we have not just provided a taxonomy of responsibility but also a notation and a general method for applying the framework to specific cases and for creating a visual representation of the responsibility relations to prompt further reflection, imagine consequences, design interventions and corrective actions, as well as consider backwards-looking liability and moral accountability.

The maritime example led to several responsibility attributions. To reiterate, the example was discussed at a relatively high degree of abstraction. Many potential attributions of liability and moral attributability have not been explored. Many actors further upstream, such as legislators, regulators and certification bodies, have been omitted to manage the size of the diagrams and the length of the discussion. But we believe the framework and notation would support more detailed discussions, by ‘zoning in’ on parts of the models in Figures 7 - 11. Attributing responsibility in cases involving AI-based systems is unlikely to be straightforward. In particular, there are many actors involved, responsible in different ways for different aspects of incidents and events. The framework, notation, and method can offer a starting point for mapping the complexity of this network of responsibilities.

There are several intended users of the framework.

First, there are the engineering and safety science communities. Safety engineering practices involve proactively determining and assessing the risk of hazards which endanger human life or damage property or the environment. Possible combinations of events leading to hazards and subsequent harms are modelled using methods such as fault trees (U.S. Nuclear Regulatory Commission, 1981) and STAMP (Leveson, 2016) to identify which potential conditions, failures or faults need to be eliminated or mitigated, and to enable assessment of the sufficiency of the mitigations regarding acceptable levels of risk. The conceptual framework and notation presented in this paper may offer a useful perspective that can enrich and help to guide safety practices (Ryan et al., 2024). By offering a lens for considering liability and moral responsibility when charting through possible scenarios in which AI system and infrastructure failure modes which could lead to unacceptable harm, it can show designers and engineers where they have a role-responsibility to ensure safety. It can also help them to evaluate whether the role-responsibilities they are placing on operators in-the-loop, such as the remote operator in the maritime scenario, are inappropriate or overly demanding and where new forms of endangerment may arise with the deployment of AI, such as individuals being liability sinks and moral crumple zones for outcomes over which they have limited control or knowledge.

Further, responsible technology development is increasingly interdisciplinary. The objectives of responsible AI innovation -- to be safe, beneficial, equitable -- require the efforts of ethicists, philosophers, lawyers, social scientists and engineers. Typically, research conversations across disciplinary boundaries start with discussions about the ways in which terms are meant. The framework provides a common language for clear discussion about responsibility attributions without the need for lengthy preliminary semantic discussions. At the same time, disciplinary experts on a particular aspect of a scenario, such as safety engineers on causation, or legal scholars on liability, might use the framework to point out to others particular areas of interest or controversy.

Second, there are corporate officials who may be considering the legal and reputational risk of decisions they must make. Though the notation presented is informal, it does help to have a visual model to guide through complex questions of responsibility in adequate detail, but still simply enough to engage all stakeholders.

Third, public officials and policy makers in the different sectors in which AI is being deployed. Just attributions of backwards-looking responsibility are particularly salient in an AI context, where the complexity and novelty of the technology, and perhaps an unreadiness of its surrounding infrastructure to support its performance, could lead to the nearest human operator being the easiest actor to hold responsible. The framework provides a way to point out where the 'liability sink' and 'moral crumple zone' problems may arise. It also offers a way to model the distribution of responsibility across a wider socio-technical system, both to shine a light on less visible responsible actors who may avoiding a responsibility attribution and to consider how actively to distribute responsibility to avoid accidents and mishaps.

Used to model forwards-looking responsibility, public officials could use the framework to consider legal duties in real and imagined cases such as the maritime example. This may allow for more specific, concrete guidance to be provided by regulators to duty-holders. UK sector-specific regulators are tasked with translating high-level principles for responsible AI into concrete guidance for AI actors across the life cycle to implement as a matter of best practice (Department for Science, Innovation & Technology, 2023). The EU's approach, as set out in the AI Act, is more prescriptive, mandating specific legal duties around conformity assessment, auditing and the monitoring of high-risk systems (European Commission, 2021; European Parliament and Council, 2024). These are the sorts of duties which could inform the requirements to which the technology corporation, vessel manufacturer and vessel owner in the scenario should have been working. The framework may also help public officials to make this guidance more operational by encouraging them to think through whether the criteria for the clarity, appropriateness, demandingness and conflicts between role-responsibilities are met.

Finally, we envisage that the framework and methodology have value for teachers and students from diverse disciplines who are interested in the attribution of responsibility for AI. By providing conceptual clarity on what 'responsibility' means, by distilling the literature in areas of philosophy and law, by providing criteria and conditions for appropriate and fair attributions of responsibility, and then presenting these in an informal graphical notation which can model complex scenarios, it can help students to engage in focused and sophisticated discussions about where responsibility does and should lie for the outputs and impacts of AI in specific cases. This will help to widen participation, bring in new perspectives and energise debates about responsible AI innovation and use.

In our experience, a perennial question in discussions around the wider implications of AI-based technologies is the question, 'who is responsible?' This question arises in part because of the novel properties that AI systems are designed to have. Amongst these properties are that they achieve goals with some degree of under-specification, which incurs uncertainty, and that we transfer decision-making functions to them, giving them the autonomy to cause harm without human intervention. Aside from the added technological complexity, the question is not straightforward to answer because of what it still leaves open. Responsible for what? Which of the many actors involved in the system lifecycle and the causal chain? In what sense of responsibility? We hope that this framework, notation, and suggested method can help discussants to consider these questions clearly and that it can be applied fruitfully to specific cases involving AI-enabled systems.

**Acknowledgements.** The work was supported by the Engineering and Physical Sciences Research Council (EP/W011239/1). The authors thank Professor Ian Pyle, Dr Laura Fearnley, and the High Integrity Systems Engineering group at the University of York for their comments on the work. We would also like to thank the anonymous reviewers, whose comments helped us considerably to improve it.

## References

Acharya, D. B., Kuppan, K., & Divya, B. (2025). Agentic AI: Autonomous intelligence for complex goals—A comprehensive survey. *IEEE Access*, *13*, 18912–18936. https://doi.org/10.1109/ACCESS.2025.3532853

Askell, A., Brundage, M., & Hadfield, G. (2019). The role of cooperation in responsible AI development. arXiv. https://doi.org/10.48550/arXiv.1907.04534

Bainbridge, L. (1983). Ironies of automation. *Automatica*, 19(6), 775–779. https://doi.org/10.1016/0005-1098(83)90046-8

Aristotle. (2011). Nicomachean ethics (R. C. Bartlett & S. D. Collins, Trans.). University of Chicago Press. https://doi.org/10.7208/chicago/9780226026763.001.0001

Beebee, H., Hitchcock, C., & Menzies, P. (Eds.). (2009). The Oxford handbook of causation. Oxford University Press. https://doi.org/10.1093/oxfordhb/9780199279739.001.0001

Björnsson, G., & Hess, K. (2017). Corporate crocodile tears? *Philosophy and Phenomenological Research*, 94(2), 273–298. https://doi.org/10.1111/phpr.12260

National Transportation Safety Board. (2019). *Collision between vehicle controlled by developmental automated driving system and pedestrian, Tempe, Arizona, March 18, 2018* (NTSB/HAR-19/03). https://www.ntsb.gov/investigations/accidentreports/reports/har1903.pdf

Braham, M., & Van Hees, M. (2009). Degrees of causation. *Erkenntnis*, 71, 323–344. https://doi.org/10.1007/s10670-009-9184-8

Braham, M., & Van Hees, M. (2012). An anatomy of moral responsibility. *Mind*, 121(483), 601–634. https://doi.org/10.1093/mind/fzs081

Burton, S., Habli, I., Lawton, T., McDermid, J., Morgan, P., & Porter, Z. (2020). Mind the gaps: Assuring the safety of autonomous systems from an engineering, ethical, and legal perspective. *Artificial Intelligence*, 279, 103201. https://doi.org/10.1016/j.artint.2019.103201

Cane, P. (2002). *Responsibility in law and morality*. Bloomsbury Publishing (Hart Publishing). https://www.bloomsbury.com/uk/responsibility-in-law-and-morality-9781847310262/

Carlson, C. S. (2012). *Effective FMEAs: Achieving safe, reliable, and economical products and processes using failure mode and effects analysis* (Vol. 1). John Wiley & Sons. https://doi.org/10.1002/9781118312575

Chan, A., Ezell, C., Kaufmann, M., Wei, K., Hammond, L., Bradley, H., Bluemke, E., Rajkumar, N., Krueger, D., Kolt, N., & Anderljung, M. (2024). Visibility into AI agents. In *Proceedings of the 2024 ACM Conference on Fairness, Accountability, and Transparency (FAccT '24)* (pp. 958–973). ACM. https://doi.org/10.1145/3630106.3658948

Chesterman, S. (2020). Artificial intelligence and the limits of legal personality. *International & Comparative Law Quarterly*, 69(4), 819–844. https://doi.org/10.1017/S0020589320000366

Cobbe, J., Veale, M., & Singh, J. (2023). Understanding accountability in algorithmic supply chains. In *Proceedings of the 2023 ACM Conference on Fairness, Accountability, and Transparency (FAccT '23)* (pp. 1186–1197). ACM. https://doi.org/10.1145/3593013.3594073

Coeckelbergh, M. (2020). Artificial intelligence, responsibility attribution, and a relational justification of explainability. *Science and Engineering Ethics*, 26(4), 2051–2068. https://doi.org/10.1007/s11948-019-00146-8

European Commission. (2021). *Proposal for a Regulation of the European Parliament and of the Council laying down harmonised rules on artificial intelligence (Artificial Intelligence Act) and amending certain Union legislative acts* (COM/2021/206 final). https://eur-lex.europa.eu/legal-content/EN/TXT/?uri=celex:52021PC0206

European Commission. (2022). *Proposal for a Directive of the European Parliament and of the Council on adapting non-contractual civil liability rules to artificial intelligence (AI Liability Directive)* (COM/2022/496 final). https://eur-lex.europa.eu/legal-content/EN/TXT/?uri=CELEX%3A52022PC0496

European Parliament and Council (2024). *Regulation (EU) 2024/1689 of the European Parliament and of the Council of 13 June 2024 laying down harmonised rules on artificial intelligence and amending Regulations (EC) No 300/2008, (EU) No 167/2013, (EU) No 168/2013, (EU) 2018/858, (EU) 2018/1139 and (EU) 2019/2144 and Directives 2014/90/EU, (EU) 2016/797 and (EU) 2017/2109.* Official Journal of the European Union, L, 2024/1689, 12.7.2024, ELI: http://data.europa.eu/eli/reg/2024/1689/oj

Law Commission and Scottish Law Commission. (2022). *Automated vehicles: Summary of joint report* (HC 1068). https://lawcom.gov.uk/project/automated-vehicles/

U.S. Nuclear Regulatory Commission. (1981). Fault Tree Handbook (NUREG-0492). https://www.nrc.gov/docs/ML1007/ML100780465.pdf

Cooper, A. F., Moss, E., Laufer, B., & Nissenbaum, H. (2022). Accountability in an algorithmic society: relationality, responsibility, and robustness in machine learning. In *Proceedings of the 2022 ACM Conference on Fairness, Accountability, and Transparency (FaccT '22)* (pp. 864–876). https://doi.org/10.1145/3531146.3533150

Copp, D. (1979). Collective actions and secondary actions. *American Philosophical Quarterly*, 16(3), 177–186. http://www.jstor.org/stable/20009757

Crootof, R., Kaminski, M. E., Price, W. N., & Nicholson, I. I. (2023). Humans in the Loop. *Vanderbilt Law Review*, 76(2), 429–510. https://wp0.vanderbilt.edu/lawreview/wp-content/uploads/sites/278/2023/03/Humans-in-the-Loop.pdf

Danks, D., Rose, D., & Machery, E. (2014). Demoralizing causation. *Philosophical Studies*, 171(2), 251–277. https://doi.org/10.1007/s11098-013-0266-8

Darwall, S. (2007). Moral obligation and accountability. In R. Shafer-Landau (Ed.), *Oxford Studies in Metaethics* (Vol. 2, pp. 111–132). Oxford University Press. https://global.oup.com/academic/product/oxford-studies-in-metaethics-9780199218073?cc=gb&lang=en&#

Davis, M. (2012). "Ain't no one here but us social forces": Constructing the professional responsibility of engineers. *Science and Engineering Ethics*, 18(1), 13–34. https://doi.org/10.1007/s11948-010-9225-3

Dekker, S. W. A., & Breakey, H. (2016). 'Just culture:' Improving safety by achieving substantive, procedural and restorative justice. *Safety Science*, 85, 187–193. https://doi.org/10.1016/j.ssci.2016.01.018

Del Frate, L., Zwart, S. D., & Kroes, P. A. (2011). Root cause as a U-turn. *Engineering Failure Analysis*, 18(2), 747–758. https://doi.org/10.1016/j.engfailanal.2010.12.006

Demirtas, H. (2022). Causation comes in degrees. *Synthese*, 200(2), 64. https://doi.org/10.1007/s11229-022-03507-2

Department for Science, Innovation & Technology. (2023). *A pro-innovation approach to AI regulation* (White Paper). https://www.gov.uk/government/publications/a-pro-innovation-approach-to-ai-regulation

Devitt, S. K. (2022). Bad, mad, and cooked: Moral responsibility for civilian harms in human-AI military teams. In J. M. Schraagen (Ed.), *Responsible use of AI in military systems*. CRC Press. https://doi.org/10.48550/arXiv.2211.06326

Dignum, V. (2019). *Responsible artificial intelligence: How to develop and use AI in a responsible way* (Vol. 2156). Springer. https://link.springer.com/book/10.1007/978-3-030-30371-6?trk=public_post_comment-text

Duff, R. A. (2007). *Answering for crime: Responsibility and liability in the criminal law*. Bloomsbury Publishing (Hart Publishing). https://www.bloomsbury.com/uk/answering-for-crime-9781849460330/

Eke, D. O., Wakunuma, K., & Akintoye, S. (2023). Introducing Responsible AI in Africa. In D. O. Eke, K. Wakunuma, & S. Akintoye (Eds.), *Responsible AI in Africa: Challenges and Opportunities* (pp. 1–11). Springer International Publishing. https://doi.org/10.1007/978-3-031-08215-3_1

Elish, M. C. (2019). Moral crumple zones: Cautionary tales in human-robot interaction. Engaging Science, Technology, and Society, 5, 40–60. https://doi.org/10.17351/ests2019.260

Endsley, M. R. (2023). Ironies of artificial intelligence. *Ergonomics*, 66(11), 1656–1668. https://doi.org/10.1080/00140139.2023.2243404

NHS England. (2022). Patient safety incident investigation. https://www.england.nhs.uk/wp-content/uploads/2022/08/B1465-PSII-overview-v1-FINAL.pdf

Fahlquist, J. N. (2015). Responsibility as a virtue and the problem of many hands. In S. van den Eede & P. K. E. Brey (Eds.), *Moral responsibility and the problem of many hands* (pp. 187–208). Routledge. https://www.taylorfrancis.com/chapters/edit/10.4324/9781315734217-7/responsibility-virtue-problem-many-hands-jessica-nihl%C3%A9n-fahlquist

Farina, M., Zhdanov, P., Karimov, A., & Lavazza, A. (2024). AI and society: A virtue ethics approach. *AI & Society*, 39(3), 1127–1140. https://doi.org/10.1007/s00146-022-01545-5

Fenelon, P., & McDermid, J. A. (1993). An integrated tool set for software safety analysis. *Journal of Systems and Software*, 21(3), 279–290. https://doi.org/10.1016/0164-1212(93)90029-W

Fischer, J. M., & Ravizza, M. (1998). *Responsibility and control: A theory of moral responsibility.* Cambridge University Press. https://doi.org/10.1017/CBO9780511814594

Fjeld, J., Achten, N., Hilligoss, H., Nagy, A., & Srikumar, M. (2020). *Principled artificial intelligence: Mapping consensus in ethical and rights-based approaches to principles for AI*. Berkman Klein Center Research Publication, 2020-1. http://dx.doi.org/10.2139/ssrn.3518482

Foot, P. (1967). The problem of abortion and the doctrine of the double effect. *Oxford Review*, (5), 1-7. https://sites.pitt.edu/~mthompso/readings/foot.pdf

Frankfurt, H. (2018). Alternate possibilities and moral responsibility. In M. McKenna & D. Widerker (Eds.), *Moral Responsibility and Alternative Possibilities* (pp. 17–25). Routledge. https://doi.org/10.4324/9781315199924

French, P. A. (1984). *Collective and corporate responsibility*. Columbia University Press. https://doi.org/10.7312/fren90672

Goetze, T. S. (2022). Mind the Gap: Autonomous Systems, the Responsibility Gap, and Moral Entanglement. In *Proceedings of the 2022 ACM Conference on Fairness, Accountability, and Transparency (FaccT '22)* (pp. 390–400). https://doi.org/10.1145/3531146.3533106

Grote, T., & Berens, P. (2020). On the ethics of algorithmic decision-making in healthcare. *Journal of Medical Ethics*, 46(3), 205–211. https://doi.org/10.1136/medethics-2019-105586
Hakli, R., & Mäkelä, P. (2019). Moral responsibility of robots and hybrid agents. *The Monist*, 102(2), 259–275. https://doi.org/10.1093/monist/onz009

Halpern, J., & Hitchcock, C. (2010). Actual causation and the art of modeling. In J. Halpern & C. Hitchcock (Eds.), *Causality, Probability, and Heuristics: A Tribute to Judea Pearl* (pp. 383–406). College Publications. https://doi.org/10.48550/arXiv.1106.2652

Hansson, S. O. (2022a). Responsibility in road traffic. In K. Edvardsson Björnberg, S. O. Hansson, M. Å. Belin, & C. Tingvall (Eds.), *The Vision Zero Handbook: Theory, Technology and Management for a Zero Casualty Policy* (pp. 1–27). Springer. https://doi.org/10.1007/978-3-030-23176-7_5-1

Hansson, S. O. (2022b). Who Is Responsible If the Car Itself Is Driving? In *Test-Driving the Future: Autonomous Vehicles and the Ethics of Technological Change* (pp. 43-58). https://doi.org/10.5771/9781786613240

Hart, H. L. A. (2008). Punishment and responsibility: *Essays in the Philosophy of Law*. Oxford University Press. https://doi.org/10.1093/acprof:oso/9780199534777.001.0001

Hart, H. L. A., & Honoré, T. (1985*). Causation in the law*. Oxford University Press. https://doi.org/10.1093/acprof:oso/9780198254744.001.0001

Himmelreich, J. (2019). Responsibility for killer robots. *Ethical Theory and Moral Practice*, 22(3), 731–747. https://doi.org/10.1007/s10677-019-10007-9

International Maritime Organization. (2024). *Convention on the International Regulations for Preventing Collisions at Sea, 1972 (COLREGs)*. Retrieved from https://www.imo.org/en/About/Conventions/Pages/COLREG.aspx

Jobin, A., Ienca, M., & Vayena, E. (2019). The global landscape of AI ethics guidelines. *Nature Machine Intelligence*, *1*(9), 389–399. https://doi.org/10.1038/s42256-019-0088-2

Johnson, D. G. (2006). Computer systems: Moral entities but not moral agents. *Ethics and Information Technology*, *8*, 195–204. https://doi.org/10.1007/s10676-006-9111-5

Kaiserman, A. (2018). 'More of a cause': Recent work on degrees of causation and responsibility. *Philosophy Compass*, *13*(7), e12498. https://doi.org/10.1111/phc3.12498

Kiener, M. (2022). Can we bridge AI's responsibility gap at will? *Ethical Theory and Moral Practice*, *25*(4), 575–593. https://doi.org/10.1007/s10677-022-10313-9

King, Z. J. (2023). What are we Praiseworthy For? In R. Chang & A. Srinivasan (Eds.), *Conversations in Philosophy, Law, and Politics*. Oxford University Press. https://doi.org/10.1093/oso/9780198864523.001.0001

Knobe, J. (2009). Folk judgments of causation. *Studies in History and Philosophy of Science Part A*, *40*(2), 238–242. https://doi.org/10.1016/j.shpsa.2009.03.009

Lawton, T., Morgan, P., Porter, Z., Hickey, S., Cunningham, A., Hughes, N., Iacovides, I., Jia, Y., Sharma, V., & Habli, I. (2024). Clinicians risk becoming 'liability sinks' for artificial intelligence. *Future Healthcare Journal*, *11*(1). https://doi.org/10.1016/j.fhj.2024.100007

Lawton, T., Porter, Z., Habli, I., Sharma, V., Morgan, P. D. J., Iacovides, J., Cunningham, A., Jia, Y., Hussain, M., Hughes, N., et al. (2025). *Avoiding the AI 'off-switch': Make AI work for clinicians, to deliver for patients*. https://eprints.whiterose.ac.uk/id/eprint/224888/

Leveson, N. G. (2016). *Engineering a safer world: Systems thinking applied to safety*. The MIT Press. https://doi.org/10.7551/mitpress/8179.001.0001

List, C., & Pettit, P. (2011). *Group agency: The possibility, design, and status of corporate agents*. Oxford University Press. https://doi.org/10.1093/acprof:oso/9780199591565.001.0001

Ludwig, K. (2017). *From plural to institutional agency: Collective action II*. Oxford University Press. https://doi.org/10.1093/oso/9780198789994.001.0001

Matthias, A. (2004). The responsibility gap: Ascribing responsibility for the actions of learning automata. *Ethics and Information Technology*, *6* (2004), 175–183. https://doi.org/10.1007/s10676-004-3422-1

McDermid, J. A., Calinescu, R., Habli, I., Hawkins, R., Jia, Y., Molloy, J., Osborne, M., Paterson, C., Porter, Z., & Ryan Conmy, P. (2024). The safety of autonomy: A systematic approach. *Computer*, *57*(4), 16–25. https://doi.org/10.1109/MC.2023.3317329

McKenna, M. (2012). *Conversation & responsibility*. Oxford University Press. https://doi.org/10.1093/acprof:oso/9780199740031.001.0001

Moore, M. S. (2009). *Causation and responsibility: An essay in law, morals, and metaphysics*. Oxford University Press. https://doi.org/10.1093/acprof:oso/9780199256860.001.0001

Morgan, P. (2023). Tort Liability and Autonomous Systems Accidents-Challenges and Future Developments. In *Tort Liability and Autonomous Systems Accidents* (pp. 1–26). Edward Elgar Publishing. https://doi.org/10.4337/9781802203844.00005

Morgan, P. (2024). Tort law and AI: Vicarious liability. In E. Lim & P. Morgan (Eds.), *The Cambridge Handbook of Private Law and Artificial Intelligence* (Chapter 6). Cambridge University Press. https://doi.org/10.1017/9781108980197

National Highway Traffic Safety Administration, Office of Defects Investigation. (2022). *Engineering Analysis 22-002. Engineering Analysis EA22-002*. https://static.nhtsa.gov/odi/inv/2022/INOA-EA22002-3184.PDF

Nissenbaum, H. (1996). Accountability in a computerized society. *Science and Engineering Ethics*, *2*, 25–42. https://doi.org/10.1007/BF02639315

Noordhof, P. (2020). *A variety of causes*. Oxford University Press. https://doi.org/10.1093/oso/9780199251469.001.0001

OECD. (2019). *AI principles: Accountability (Principle 1.5)*. https://www.oecd.org/en/topics/ai-principles.html

Okolo, C. T., Aruleba, K., & Obaido, G. (2023). Responsible AI in Africa—Challenges and Opportunities. *Responsible AI in Africa: Challenges and Opportunities*, 35–64. https://doi.org/10.1007/978-3-031-08215-3_3

Owen, R., Stilgoe, J., Macnaghten, P., Gorman, M., Fisher, E., & Guston, D. (2013). A framework for responsible innovation. In *Responsible innovation: managing the responsible emergence of science and innovation in society* (pp. 27–50). https://doi.org/10.1002/9781118551424.ch2

Pearl, J. (2009). *Causality*. Cambridge University Press. https://doi.org/10.1017/CBO9780511803161

Percy, R. A., & Charlesworth, J. (2018). *Charlesworth & Percy on Negligence*. Sweet & Maxwell. https://heinonline.org/HOL/LandingPage?handle=hein.journals/tortllr1&div=19&id=&page=

Pettit, P. (2007). Responsibility incorporated. *Ethics*, *117*(2), 171–201. https://doi.org/10.1086/510695

Porter, Z., Habli, I., Monkhouse, H., & Bragg, J. (2018). The moral responsibility gap and the increasing autonomy of systems. In B. Gallina, A. Skavhaug, E. Schoitsch, & F. Bitsch (Eds.), Computer Safety, Reliability, and Security. SAFECOMP 2018. Lecture Notes in Computer Science (Vol. 11094). Springer, Cham. https://doi.org/10.1007/978-3-319-99229-7_43

Porter, Z., Zimmermann, A., Morgan, P., McDermid, J., Lawton, T., & Habli, I. (2022). Distinguishing two features of accountability for AI technologies. *Nature Machine Intelligence*, *4*(9), 734–736. https://doi.org/10.1038/s42256-022-00533-0

Prifti, K., Stamhuis, E., & Heine, K. (2022). Digging into the Accountability Gap: Operator's Civil Liability in Healthcare AI-systems. In *Law and Artificial Intelligence: Regulating AI and Applying AI in Legal Practice* (pp. 279–295). Springer. https://doi.org/10.1007/978-94-6265-523-2_15

Purves, D., Jenkins, R., & Strawser, B. J. (2015). Autonomous machines, moral judgment, and acting for the right reasons. *Ethical Theory and Moral Practice*, *18*, 851–872. https://doi.org/10.1007/s10677-015-9563-y

Rubel, A., Castro, C., & Pham, A. (2019). Agency laundering and information technologies. *Ethical Theory and Moral Practice*, *22*, 1017–1041. https://doi.org/10.1007/s10677-019-10030-w

Ryan Conmy, P., McDermid, J., Habli, I., & Porter, Z. (2023). Safety engineering, role responsibility and lessons from the Uber ATG Tempe Accident. In *Proceedings of the First International Symposium on Trustworthy Autonomous Systems (TAS '23)* (pp. 1–10). ACM. https://doi.org/10.1145/3597512.3599718

Ryan, P., Habli, I., Al-Qaddoumi, J., Porter, Z., & McDermid, J. (2024). *What's my role? Modelling responsibility for AI-based safety-critical systems*. arXiv. https://arxiv.org/abs/2401.09459

Ryan, P., Porter, Z., Fearnley, L. C. A., McDermid, J. A., Habli, I., Al-Qaddoumi, J., & Morgan, P. D. J. (2025). *Evidencing responsibility for use of AI in safety critical systems.* https://eprints.whiterose.ac.uk/id/eprint/224048/

Robichaud, P., & Wieland, J. W. (2017). *Responsibility: The epistemic condition*. Oxford University Press. https://doi.org/10.1093/oso/9780198779667.001.0001

Santoni de Sio, F., & Van den Hoven, J. (2018). Meaningful human control over autonomous systems: A philosophical account. *Frontiers in Robotics and AI*, *5*, 15. https://doi.org/10.3389/frobt.2018.00015

Scanlon, T. M. (2000). *What we owe to each other*. Harvard University Press. https://www.hup.harvard.edu/books/9780674004238

Schiff, D., Rakova, B., Ayesh, A., Fanti, A., & Lennon, M. (2020). *Principles to practices for responsible AI: closing the gap*. https://arxiv.org/abs/2006.04707

Sebastián, M. Á. (2021). First-person representations and responsible agency in AI. *Synthese*, *199*(3), 7061–7079. https://doi.org/10.1007/s11229-021-03105-8

Sepinwall, A. J. (2016). Corporate moral responsibility. *Philosophy Compass*, *11*(1), 3–13. https://doi.org/10.1111/phc3.12293

Shoemaker, D. (2011). Attributability, answerability, and accountability: Toward a wider theory of moral responsibility. *Ethics*, *121*(3), 602–632. https://doi.org/10.1086/659003

Shoemaker, D. (2013). On criminal and moral responsibility. *Oxford Studies in Normative Ethics*, *3*, 154–178. https://doi.org/10.1093/acprof:oso/9780199685905.003.0008

Shoemaker, D. (2015). *Responsibility from the Margins*. Oxford University Press. https://doi.org/10.1093/acprof:oso/9780198715672.001.0001

Smith, H. (2021). Artificial intelligence to inform clinical decision making: A practical solution to an ethical and legal challenge. In *IJCAI 2020 AI for Social Good workshop*. IJCAI. https://projects.iq.harvard.edu/sites/projects.iq.harvard.edu/files/crcs/files/ai4sg-21_paper_22.pdf

Stahl, B. C. (2023). Embedding responsibility in intelligent systems: from AI ethics to responsible AI ecosystems. *Scientific Reports*, *13*(1), 7586. https://doi.org/10.1038/s41598-023-34622-w

Strawson, P. F. (2008). *Freedom and resentment and other essays*. Routledge. https://doi.org/10.4324/9780203882566

Stuart, M. T., & Kneer, M. (2021). Guilty artificial minds: Folk attributions of mens rea and culpability to artificially intelligent agents. *Proceedings of the ACM on Human-Computer Interaction*, *5*(CSCW2), 1–27. https://doi.org/10.1145/34795

Tadros, V. (2018). Causal contributions and liability. *Ethics*, *128*(2), 402–431. https://doi.org/10.1086/694275

Talbert, M. (2016). *Moral Responsibility: An Introduction*. Polity Press. https://www.politybooks.com/bookdetail?book_slug=moral-responsibility-an-introduction--9780745680583

Ten Holter, C. (2022). Participatory design: lessons and directions for responsible research and innovation. *Journal of Responsible Innovation*, *9*(2), 275–290. https://doi.org/10.1080/23299460.2022.2041801

Thompson, D. F. (1980). Moral responsibility of public officials: The problem of many hands. *American Political Science Review*, *74*(4), 905–916. https://doi.org/10.2307/1954312

Tigard, D. W. (2021). There is no techno-responsibility gap. *Philosophy & Technology*, *34*(3), 589–607. https://doi.org/10.1007/s13347-020-00414-7

Tobia, K., Nielsen, A., & Stremitzer, A. (2021). When does physician use of AI increase liability? *Journal of Nuclear Medicine*, *62*(1), 17–21. https://doi.org/10.2967/jnumed.120.256032

UNESCO. (2021). *Recommendation on the Ethics of Artificial Intelligence* (SHS/BIO/PI/2021/1). https://unesdoc.unesco.org/ark:/48223/pf0000381137

Stanford University. (2023). *AI Index Report 2023*. https://aiindex.stanford.edu/report/

Unruh, C. F. (2022). Doing and allowing good. *Analysis*, *82*(4), 630–637. https://doi.org/10.1093/analys/anac018

Vallor, S. (2016). *Technology and the Virtues: A Philosophical Guide to a Future Worth Wanting*. Oxford University Press. https://doi.org/10.1093/acprof:oso/9780190498511.001.0001

Vallor, S., & Ganesh, B. (2023). Artificial intelligence and the imperative of responsibility: Reconceiving AI governance as social care. In *The Routledge Handbook of Philosophy of Responsibility* (pp. 395–406). Routledge. https://doi.org/10.4324/9781003282242

Van de Poel, I. (2011). The relation between forward-looking and backward-looking responsibility. In *Moral responsibility: Beyond free will and determinism* (pp. 37–52). Springer. https://doi.org/10.1007/978-94-007-1878-4_3

Van de Poel, I., Royakkers, L., & Zwart, S. D. (2015). *Moral responsibility and the problem of many hands*. Routledge. https://doi.org/10.4324/9781315734217

Van de Poel, I., & Sand, M. (2021). Varieties of responsibility: two problems of responsible innovation. *Synthese*, *198* (Suppl 19), 4769–4787. https://doi.org/10.1007/s11229-018-01951-7

Véliz, C. (2021). Moral zombies: why algorithms are not moral agents. *AI & Society*, *36*, 487–497. https://doi.org/10.1007/s00146-021-01189-x

Verdicchio, M., & Perin, A. (2022). When doctors and AI interact: On human responsibility for artificial risks. *Philosophy & Technology*, *35*(11). https://doi.org/10.1007/s13347-022-00506-6

Von Schomberg, R. (2013). A vision of responsible research and innovation. In *Responsible innovation: Managing the responsible emergence of science and innovation in society* (pp. 51–74). https://doi.org/10.1002/9781118551424.ch3

Watson, G. (1996). Two faces of responsibility. *Philosophical Topics*, *24*(2), 227–248. http://www.jstor.org/stable/43154245

Williams, G. (2008). Responsibility as a virtue. *Ethical Theory and Moral Practice*, *11*(4), 455–470. https://doi.org/10.1007/s10677-008-9109-7

Wittgenstein, L. (1976). Cause and effect: Intuitive awareness. *Philosophia*, *6* (3-4), 409–425. https://doi.org/10.1007/BF02379281

Wright, R. W. (1985). Causation in tort law. *California Law Review*, *73*, 1735. https://doi.org/10.2307/3480373

Zerilli, J. (2021). *A citizen's guide to artificial intelligence*. The MIT Press. https://doi.org/10.7551/mitpress/12518.001.0001